\documentclass[runningheads]{llncs}
\usepackage{tikz}
\usepackage{graphicx}
\usepackage{comment}
\usepackage{amsmath,amssymb} 
\usepackage{color}
\usepackage{mcite}
\usepackage{subcaption}
\usepackage{multicol}
\usepackage{multirow}
\usepackage{algorithm}
\usepackage{breakcites}
\usepackage{algorithmic}
\usepackage[width=122mm,left=12mm,paperwidth=146mm,height=193mm,top=12mm,paperheight=217mm]{geometry}
\usepackage[pagebackref=true,breaklinks=true,letterpaper=true,colorlinks,bookmarks=false]{hyperref}
\makeatletter
\newcommand{\thickhline}{%
    \noalign {\ifnum 0=`}\fi \hrule height 1pt
    \futurelet \reserved@a \@xhline
}
\makeatother

\makeatletter
\newcommand{\printfnsymbol}[1]{%
  \textsuperscript{\@fnsymbol{#1}}%
}
\makeatother

\newcommand{\eg}{\textit{e}.\textit{g}.}

\begin{document}
\pagestyle{headings}
\mainmatter
\def\ECCVSubNumber{6936}  

\title{Feature Space Augmentation for Long-Tailed Data} 


\titlerunning{Feature Space Augmentation for Long-Tailed Data}
%
\author{
Peng Chu \inst{1}\thanks{Work was done at GE Research.} \and
Xiao Bian\inst{2}\printfnsymbol{1} \and
Shaopeng Liu\inst{3} \and
Haibin Ling\inst{4,1}
}
\authorrunning{P. Chu et al.}
%
\institute{
Temple University, USA \email{pchu@temple.edu}
\and
Google Inc., USA \email{xbian@google.com}
\and
GE Research, USA \email{sliu@ge.com}
\and
Stony Brook University, USA \email{hling@cs.stonybrook.edu}
}
\maketitle
\begin{abstract}
Real-world data often follow a long-tailed distribution as the frequency of each class is typically different. For example, a dataset can have a large number of under-represented classes and a few classes with more than sufficient data. However, a model to represent the dataset is usually expected to have reasonably homogeneous performances across classes. Introducing class-balanced loss and advanced methods on data re-sampling and augmentation are among the best practices to alleviate the data imbalance problem. However, the other part of the problem about the under-represented classes will have to rely on additional knowledge to recover the missing information. 

In this work, we present a novel approach to address the long-tailed problem by augmenting the under-represented classes in the feature space with the features learned from the classes with ample samples. In particular, we decompose the features of each class into a class-generic component and a class-specific component using class activation maps. Novel samples of under-represented classes are then generated on the fly during training stages by fusing the class-specific features from the under-represented classes with the class-generic features from confusing classes. 
Our results on different datasets such as iNaturalist, ImageNet-LT, Places-LT and a long-tailed version of CIFAR have shown the state of the art performances. 
\end{abstract}

\begin{figure}
\begin{minipage}[t]{.56\linewidth}
	\centering
	\includegraphics[width=1\linewidth]{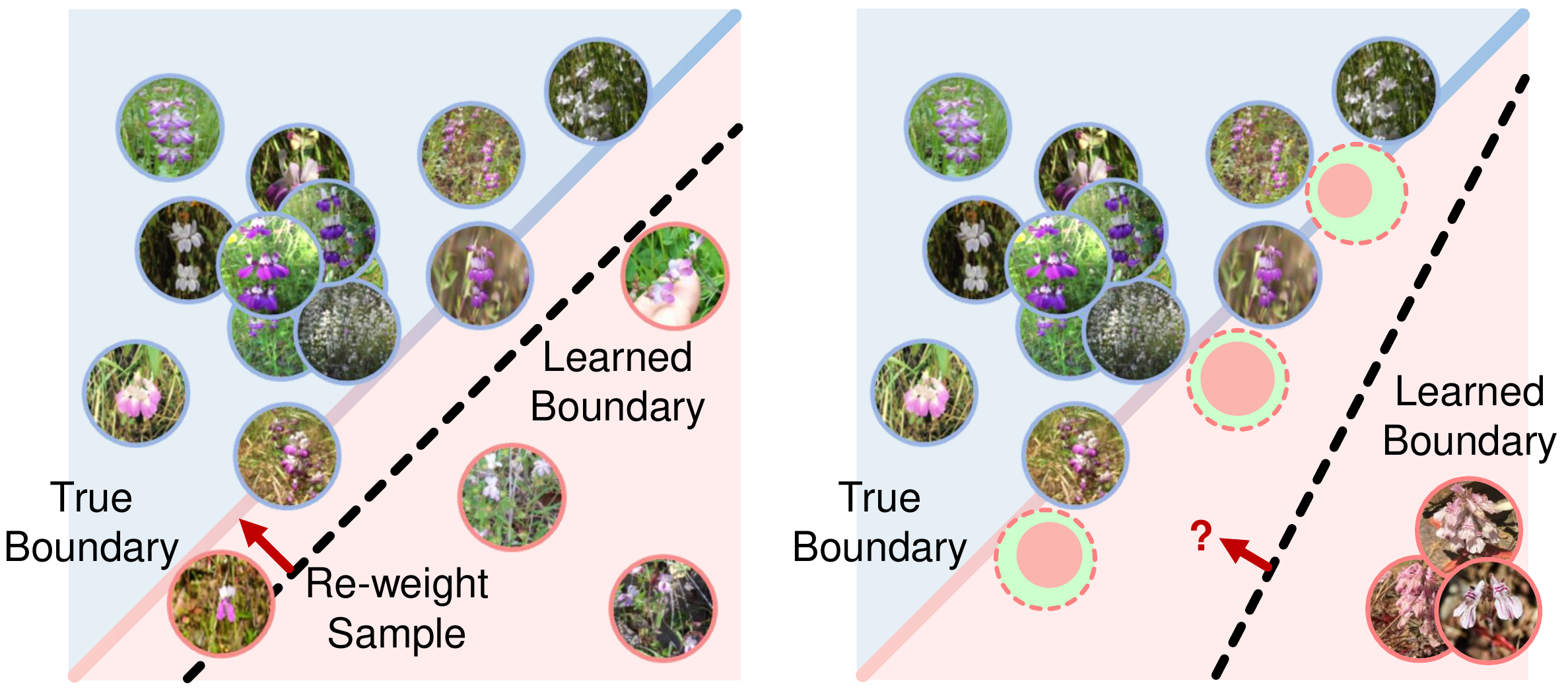}
	\caption{Left: With limited but well-spread data, the optimal decision boundary search can be recovered by sample re-weighting/loss balancing. Right: Without sufficient sample coverage, the ``optimal direction" to move the decision boundary becomes unclear. In this paper, augmented samples are generated to recover the underlying distribution.}
	\label{fig:teaser}
	\end{minipage}
	\hfill
	\begin{minipage}{.41\linewidth}
	    \centering
        \includegraphics[width=1\linewidth]{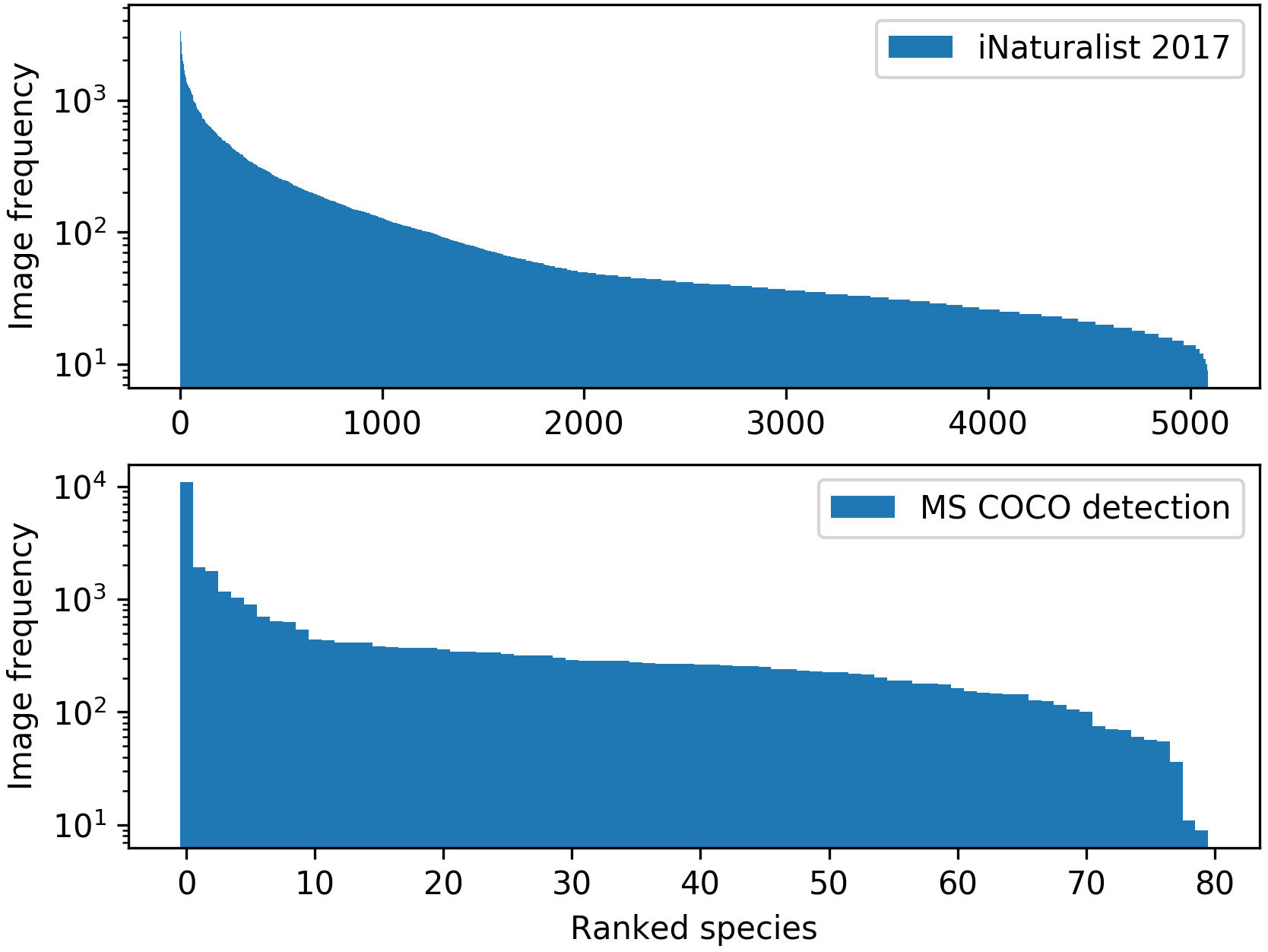}
        \caption{The sorted sample size of each class from different dataset follows similar long-tailed distribution.}
        \label{fig: longtail}
	\end{minipage}
\end{figure}

\section{Introduction}
Deep neural networks have shown considerable success in a wide variety of visual recognition tasks. Its effectiveness and generalizability have been well proved by many state-of-the-art work~\cite{krizhevsky2012imagenet, he2016deep, girshick2015fast, long2015fully} and a wide variety of real-world applications in different industries~\cite{chen2015deepdriving, zhu2018visdrone, bian2016multiscale, esteva2019guide}. However, there is often  one underlying condition that each category of interest needs to be well represented. 

To quantify the ``representativeness" of data can be a challenging problem itself. In practice, it is usually scrutinized using different heuristics, and one common criterion could be the balance of a dataset. Indeed, many public datasets are intentionally organized to have the same number of samples from each class~\cite{ILSVRC15, krizhevsky2009learning}. For problems such as segmentation and detection which are hard to ensure exact balanced data, it is always preferable to ensure good data coverage in a way that the rare classes still have sufficient data and are hence well represented~\cite{van2018inaturalist, everingham2010pascal}. 

However, real-world visual understanding problems are often fine-grained and long-tailed. To achieve human-level visual understanding, it almost implies the ability to distinguish the subtle differences between fine-grained categories and to robustly handle the presence of rare categories~\cite{akata2015evaluation}. In fact, these two properties of real-world data usually accompany each other as a large number of fine-grained categories often leads to a highly imbalanced dataset, as illustrated in Fig.~\ref{fig: longtail}. For example, in iNaturalist dataset 2017~\cite{van2018inaturalist} for species classification, there are a total of 5089 classes with the largest classes more than 1000 samples and the smallest classes fewer than 10. In iNaturalist competition 2019, even with an effort to filter out species that have insufficient observations and to further cap the maximum class size to be 500, there is still serious imbalance in the dataset as the smallest classes around 10 samples. Similar data distribution can be observed in other applications, such as a UAV-based object detection dataset~\cite{zhu2018visdrone} and COCO~\cite{lin2014microsoft}. 

Like many supervised learning algorithms, the performance of deep neural networks also suffers when the training data is highly imbalanced~\cite{cui2019class}. The problem can get worse when the categories with fewer data are severely under-sampled to the extent that the variation within each category is not fully captured by the given data~\cite{wang2018low, bengio2015sharing, yin2018feature}.

The common presence of long-tailed data in real-world problems has led to several effective practices to achieve an overall performance improvement of a given machine learning model. For example, \textit{data manipulation} such as augmentation, under-sampling and over-sampling~\cite{buda2018systematic, geifman2017deep, drummond2003c4}, and \textit{balanced loss function design} (\eg, focal-loss~\cite{lin2017focal} and class-balanced loss~\cite{cui2019class}), are the two mainstream approaches. These practices often improve the performance reasonably yet the improvement deteriorates when certain categories are severely under-represented, as shown in Fig.~\ref{fig:teaser}. Specifically, these methods are often designed to move the class decision boundary to reduce the bias introduced by imbalanced classes. However, when a class is severely under-represented such that it is hard to draw its complete data distribution, finding the right direction to adjust the decision boundary becomes challenging. We therefore focus on exploring the information learned from the head classes (the ones with ample samples) to help the tail classes (the under-represented ones) in a long-tailed dataset.

In this work, we present a novel method to address the long-tailed data classification problem by augmenting the tail classes in the feature space using the information from the head classes. In particular, we insert an attention unit with the help of the class activation map (CAM)~\cite{zhou2016learning} to filter out class-specific features and class-generic features from each class. For the samples of each tail class, we augment the high-level features (from high-level deep network layers) by mixing its class-specific features with the class-generic features from the head classes. This method is based on two underlying assumptions: 1) information from the head classes, represented as class-generic features, can help to recover the distribution of tail classes; and 2) the class-generic and class-specific features can be extracted and re-mixed to generate novel samples in the high-level feature space due to a more ``linear" representation at that level. We have designed an end-to-end training pipeline to efficiently perform such feature space augmentation, and evaluated our method on artificially created long-tailed CIFAR-10 and CIFAR-100 datasets~\cite{krizhevsky2009learning}, ImageNet-LT, Places-LT~\cite{liu2019large} and naturally long-tailed datasets such as iNaturalist 2017 \& 2018~\cite{van2018inaturalist}. Our approach has shown the state of the art performance on these long-tailed datasets compared to other mainstream deep learning models on data imbalance problems. 

\section{Related Work}
In this section, we first discuss the two directly related approaches, \textit{learning with balanced loss} and \textit{data augmentation}, and then discuss the difference and relation of our approach to few-shot learning and transfer learning. 

\subsection{Learning with Balanced Loss}
\label{balance_loss}
One of the most common and often most effective practices is \textit{learning with balanced loss}. The key to such approaches is to counter the effect of a skewed data distribution by adjusting the weights of the samples from the small classes in the loss function. It is typically accomplished by: 1) over-sampling or under-sampling~\cite{shen2016relay, buda2018systematic, zou2018unsupervised, geifman2017deep, kang2019decoupling} to achieve an even data distribution across various classes, and/or 2) assigning proper weights to the loss terms corresponding to the tail classes~\cite{ting2000comparative, huang2016learning, khan2017cost, lin2017focal, cui2019class, zhou2005training,  sarafianos2018deep, elkan2001foundations, zhang2017range}. 

Specifically, these approaches treat the issue of long-tailed data as an optimization problem such that an ``optimal" classification boundary can be recovered by carefully adjusting the weight/frequency of each data point in the training set. They typically have the advantage of a relatively clean implementation by either adjusting the loss function \cite{cui2019class, lin2017focal} or manipulating the input batch \cite{ geifman2017deep, shen2016relay, buda2018systematic, zou2018unsupervised}, and hence were widely adopted in practice \cite{shen2016relay, lin2017focal, zou2018unsupervised}.  However, when the samples of the tail classes are far from sufficient to recover the true distribution, the performance of such methods deteriorates \cite{cui2019class}. 

Two works along this direction, class-balanced loss \cite{cui2019class} and focal loss \cite{lin2017focal} draw our attention in particular for their generic applications on deep learning models. Specifically, focal loss weights the loss term of each sample based on the probability generated from the last soft-max layer \cite{lin2017focal}. It implicitly gives higher weights to samples from the tail classes to counter the bias introduced by the sample size. In~\cite{cui2019class}, the concept of the effective number is introduced to calculate the weight of each class in the loss term. 

Note that our approach can be used jointly with approaches such as focal loss~\cite{lin2017focal} and potentially gain the benefits from both. For example, we can use the feature space augmentation approach to facilitate the performance of the tail classes, and at the same time, balanced loss methods such as focal loss can give higher weights to the hard examples regardless of the class label during training.

\subsection{Data Synthesis and Augmentation}
\label{data_aug}
Generating and synthesizing new samples to compensate the small sample size of a tail class is a natural way to improve the performance of deep learning models on long-tailed data. These samples can be either generated from similar samples~\cite{chawla2002smote} or synthesized based on the given information of a dataset~\cite{he2008adasyn, zou2018unsupervised, gidaris2018dynamic}. The general application of data augmentation in different deep learning models also boosts the interest of developing more sophisticated data augmentation methods. 
For example, in~\cite{chen2019destruction}, an input image is partitioned into local regions which are shuffled during training. In~\cite{chen2019image}, local regions of an image are replaced by unlabeled data to generate synthetic images to help training. In~\cite{wang2018low}, a parametric generative model takes noise and existed samples to hallucinate new samples to support training. 

As directly manipulating the raw input images may as well introduce unexpected noise, feature vectors are instead generated by training a function to learn the relation between a pair of samples from one class and applies it to the samples in another \cite{hariharan2017low}. 
Furthermore, recent progress in generative adversarial networks (GAN) have inspired advanced methods using generative models to address the data insufficiency problem~\cite{xian2018feature}. 

In contrast to the existing approaches on augmenting feature vectors, we focus on modeling the feature space itself rather than training a heavily parametric model that applies to all different classes. The decomposition of feature space is then used to formulate novel training samples in the feature space on the fly.

\subsection{Transfer Learning}
\label{transfer_learning}
Past works in the domain of transfer learning and few-shot learning~\cite{wang2017learning, bengio2015sharing, ouyang2016factors, yin2018feature, oh2016deep, zhou2020bbn} have been conducted to solve the long-tailed problem. Our work shares a similar assumption with these works that the information from the head classes can be used to help the tail classes. However, we explicitly distinguish the generic features and specific features from each class instead of making strong assumptions on the general transferability of knowledge from the head classes to the tail classes. Specifically, in \cite{wang2017learning}, a meta-network is trained to predict the many-shot parameters from few-shot model parameters using data from the head classes with the assumption that the model parameters from different classes share a similar dynamic behavior even if the size of the training set varies. In~\cite{bengio2015sharing}, the representation is shared in general with different embedding approaches across different classes. In \cite{yin2018feature}, the variance of the head classes is learned and transferred to the tail classes with the underlying assumption that each class has its own mean but a shared variance. 
In \cite{ouyang2016factors}, visually similar classes are clustered together in order to reduce the level of data imbalance. Knowledge can then be transferred from each cluster to its sub-classes during the fine tuning stage of deep networks for object detection specifically.  

In comparison, we intentionally separate the features of each class into class-specific features and class-generic features. Only class-generic features from head classes are seen as transferable knowledge and are hence used for feature space augmentation on the tail classes.

\section{The Problem of Long Tail}
In this section, we first analyze the underlying issues of long-tailed data that affect model performance (Sec. \ref{longTail1}), and then explore deeper into the feature space of DNNs and illustrate a novel way to alleviate the problem (Sec. \ref{longTail2}). 

\subsection{Two Reasons of Model Performance Drop}
\label{longTail1}
Long-tailed data hurt the performance of learning-based classification models mainly due to the following two issues: (1) data imbalance which is relatively easy to solve, and (2) missing coverage of the data distribution caused by limited data, which is harder to deal with. 

The data imbalance issue has been discussed in several recent works~\cite{cui2019class, buda2018systematic, geifman2017deep, lin2017focal} with good solutions proposed to minimize its impact. This problem is essentially about the bias introduced by the different number of samples in the dataset. With carefully designed sampling schemes and/or loss weights, we can compensate this negative impact and move the classification decision boundary in the right direction. For example, a common practice of training on an imbalanced dataset is to over-sample the small classes or under-sample the large classes~\cite{drummond2003c4}. This is built upon the assumption that the underlying decision boundary is indeed well-defined with the given data, and hence with careful adjustment we can find its optimal location. 

\begin{figure}
    \begin{minipage}[t]{.47\linewidth}
	    \centering
	    \includegraphics[width=1\linewidth]{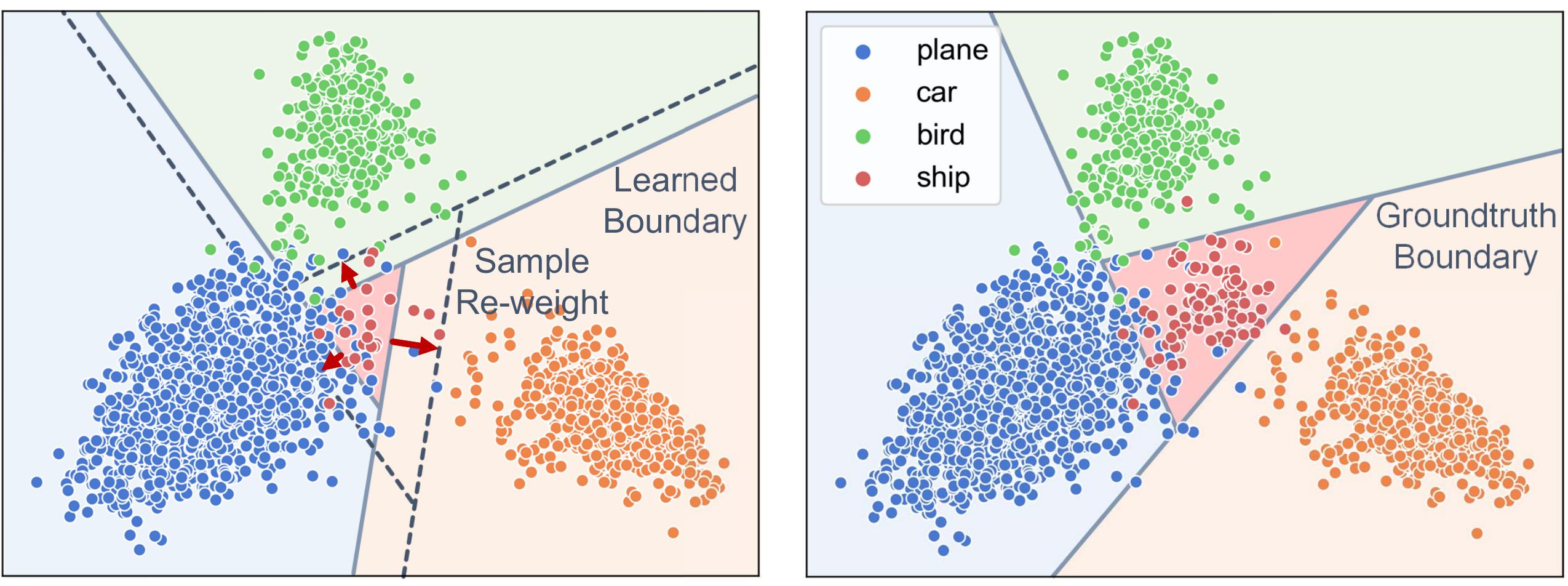}
	   \caption{The difference between the two ``optimal" decision boundaries.}
	    \label{fig:perf-drop}
	\end{minipage}
	\hfill
	\begin{minipage}[t]{.47\linewidth}
		\centering
	    \includegraphics[width=1\linewidth]{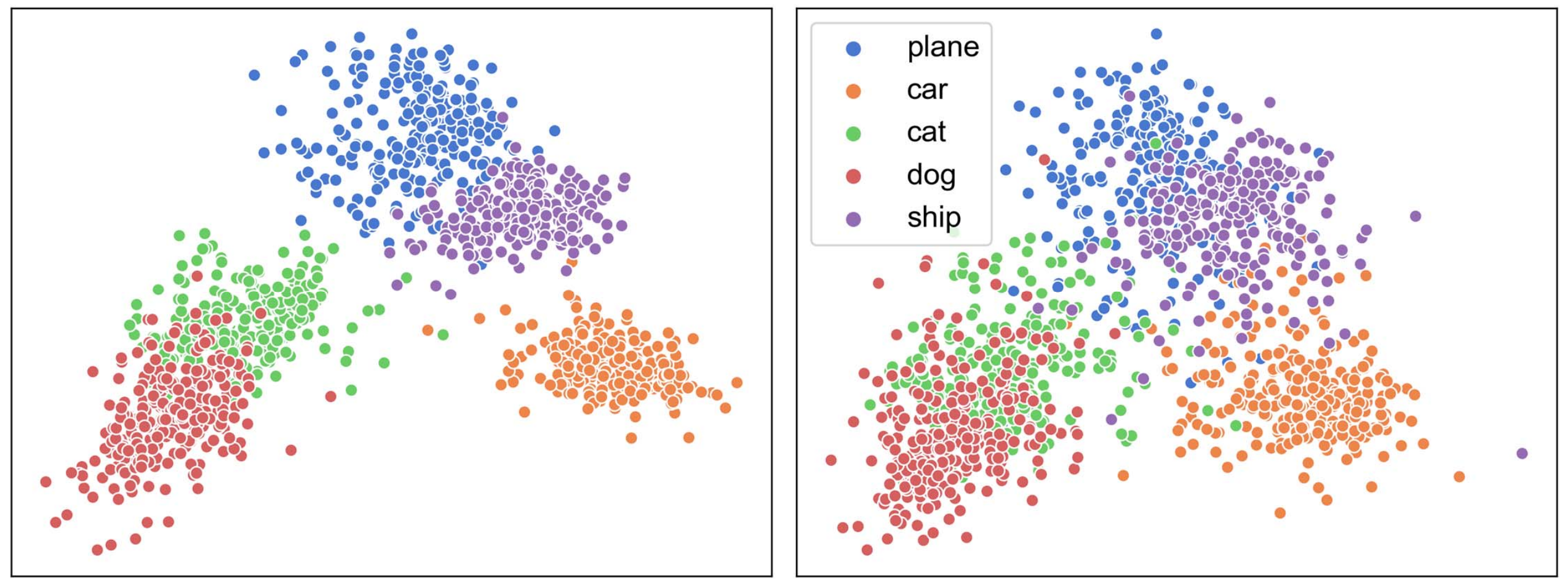}
	   \caption{Left: class-specific features, Right: class-generic features.}
	    \label{fig:fg-bg}
	\end{minipage}
\end{figure}

However, when there is simply no sufficient data for the tail classes to recover their underlying distribution, the problem of finding an optimal decision boundary becomes ill-defined. In this scenario, it becomes extremely difficult to guess the location of the decision boundary without recovering the distribution first. We hypothesize that the knowledge obtained from the head classes can help with solving the issue. 

We further elaborate the issue in Fig.~\ref{fig:perf-drop} by plotting the feature distribution of 4 classes in CIFAR-10. The features are from the last fully-connected (FC) layer of ResNet-18 and then embedded in 2-D space. When the ship class is under-represented, as shown in the left graph, simply moving the decision boundary will not provide the optimal decision boundary (as shown in the right graph) as if there were sufficient samples. 

\subsection{Class Activation Map and Feature Decomposition}
\label{longTail2}

With limited data in the tail classes and ample data in the head classes, it seems natural to use the knowledge learned from the head classes to help recovering the missing information in the tail classes. However, we have to be careful to differentiate the class-generic information that can be used to recover the distribution of the tail classes from the class-specific information that may mislead the recovery of the distribution of the tail classes. 

Inspired by the recent works on attention and visual explanation \cite{zhou2016learning, selvaraju2017grad}, we find that deep neural network features can be decomposed into two such components in a similar fashion. In particular, let us define class activation map $M_c$ of class $c$ as in \cite{zhou2016learning},
\begin{align}
M_c(x, y) = \sum_k w_k^c f_k(x, y),
\end{align}
where $f_k(x, y)$ is the feature vector in location $(x, y)$ of channel $k$, and $w_k^c$ the weights of the last layer of classifier corresponding to class $c$. The larger value of $M_c(x, y)$, the more important of feature vector at $(x, y)$ is to class $c$, and vice versa. 

We further normalize the value of $M_c(x, y)$ to the range of 0 and 1. Therefore, given a pair of thresholds $0 < \tau_s, \tau_g < 1$, we can decompose the class activation map $M_c$ into two parts, $M_c^s$ and $M_c^g$, to separate the feature vectors into class-specific features and class-generic features as follows, 
\begin{align}
\label{eq:mask}
    M_c^s &= \text{sgn}(M_c - \tau_s) \odot M_c, \\
    M_c^g &= \text{sgn}(\tau_g - M_c) \odot M_c,
\end{align}
where $\odot$ is the Hadamard product between two tensors, $\text{sgn}(x) = 1$ for $x \ge 0$ and $\text{sgn}(x) = 0$ for $x < 0$. 

Fig.~\ref{fig:fg-bg} shows the scatter plot of class-generic features and class-specific features of different classes from CIFAR-10 (More results can be seen in the supplemental material). We can see that after decomposition, even when embedded in a 2-D space, the class-specific features are clearly more separated than class-generic features. In general, we have observed a much stronger correlation between class-generic features than class-specific features across different classes and different datasets. These results further substantiate our approach on using class-generic features to augment the tail classes during training. 

\begin{figure*}[t]
	\centering
	\includegraphics[width=0.995\linewidth]{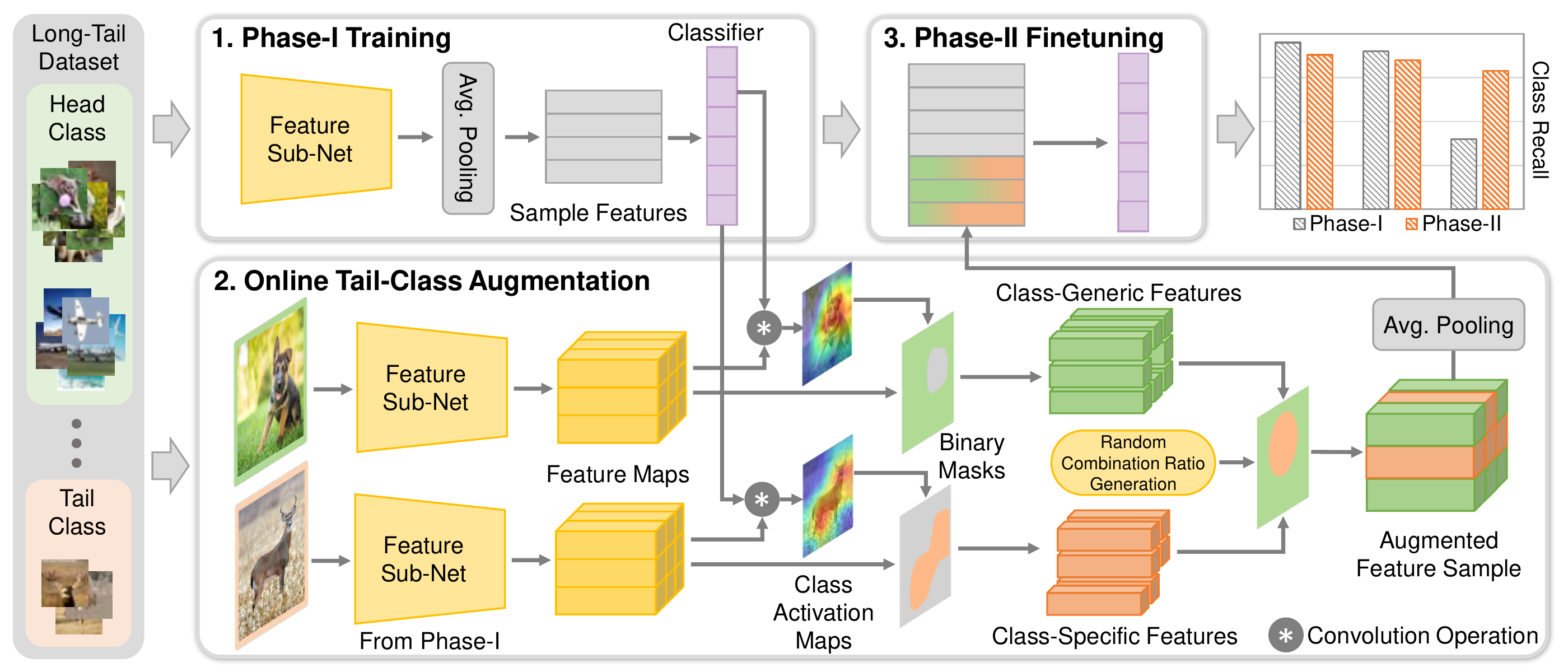}
	\caption{Overview of the proposed two-phase learning scheme.}
	\label{fig:overview}
	
\end{figure*}

\section{Method}

We propose a two-phase training scheme to leverage the class-generic information to recover the distribution of tail classes, as shown in Fig.~\ref{fig:overview}. In Phase-I, samples from all classes are used to learn the feature representation and a base classifier. In Phase-II, online feature space augmentation is applied to generate novel samples for tail classes.

\subsection{Initial Feature Learning}

In the Phase-I training, we use all images in the dataset to learn the feature sub-network and the base classifier. In order to calculate the class activation maps in the following steps, we choose a network architecture that contains a single FC layer as the final classifier, which takes input from a global average pooling layer as illustrated in Fig.~\ref{fig:overview}. A number of the modern deep convolutional neural network architectures fit into this category, e.g., ResNet~\cite{he2016deep}, DenseNet~\cite{huang2017densely}, MobileNet~\cite{sandler2018mobilenetv2}, and EfficientNet~\cite{tan2019efficientnet}.

\subsection{Feature Space Augmentation}

With the pre-trained feature sub-network and the classifier, augmented samples can be generated in the feature space on the fly by mixing the class-specific features from the given tail class and the class-generic features. 

One question we need to address is that, given a tail class, how to choose the classes from which the class-generic features will be extracted. A naive solution would be to randomly select the classes from the training dataset. However, the class-generic features from different classes may vary with each other, and such features of a randomly selected class cannot always guarantee a good recovery of the classification decision boundary. From the perspective of the optimal classification decision boundary, we observe that the ``nearby" classes in the feature space, i.e. the most ``confusing" classes with respect to the given tail class, have the biggest impact on recovering the previously ill-defined decision boundary, as seen in Fig.~\ref{fig:perf-drop}. Specifically, we calculate the classification scores for all other classes for each training sample in a given tail class, and then find its top $N_f$ confusing classes by ranking the average classification scores of other classes over all samples within the tail class.


As described in Sec.~\ref{longTail2}, we use class activation maps to separate the class-generic and class-specific information from a given image. As shown in Fig.~\ref{fig:overview}, the feature sub-network trained in Phase-I is used to extract feature maps for each input image. The weights of the linear classifier trained in Phase-I are adapted to form the $1 \times 1$ convolutional filter for each class. For each input image, the filter associated with the ground truth class is applied on its feature maps to generate the class activation map which is further normalized to the range of $[0, 1]$ for consistency. Two independent thresholds $\tau_g$ and $\tau_s$  are used to extract the corresponding binarized masks for class-generic features and class-specific features following Eq.~\ref{eq:mask}.

The class-generic information in the confusing classes is then leveraged to generate the augmented samples of each tail class in order to recover its intrinsic data distribution. 
Directly blending information at the pixel level often introduces artificial edges and hence imposes bias to the augmented samples. We, therefore, conduct the fusion in the feature space to suppress the noise and potential bias. In particular, for each real sample in the tail class, we sample $N_a$ images from its $N_f$ confusing classes. The class-specific features from the sample are then combined with the class-generic features from the $N_a$ samples in a linear way. A random combination ratio is generated to guide the fusion by randomly drawing class-generic and class-specific feature vectors to form an augmented sample for the tail class. By randomly modulating the combination ratio between the class-generic and class-specific features, the sample variance is built into this augmentation procedure. In the end, a total of $N_a$ augmented samples are generated for each real sample from the tail class. 



\subsection{Fine Tuning with Online Augmented Samples}
\label{sec:online_aug}

\begin{algorithm} [!t]
	\small
	\caption{\small Online Feature Augmentation}
	\begin{algorithmic}[1]
		\STATE \textbf{Input}: All training images features $\mathbb{F}$ and their CAM $\mathbb{M}$. \\
		\STATE \textbf{Output}: Training batch with augmented feature samples $b_{out}$.\\
        \STATE Initialize output batch $b_{out}$.
        \FOR {$i = 1,\dots, N_t$}
            \STATE $\textbf{F}_c$ $\gets$  Draw one sample from tail classes 
            \STATE Append $\textbf{F}_c$ to $b_{out}$
            \STATE $M_c^s \gets M_c > \tau_s$
            \STATE $\textbf{F}_c^s \gets M_c^s \odot \textbf{F}_c$
            \STATE $\{u\}$ $\gets$ Find confusing classes for class $c$
            \FOR {$u$ in $\{u\}$}
                \STATE $\textbf{F}_u$ $\gets$ Draw one sample from class $u$ 
                \STATE $M_u^g \gets M_u < \tau_g$
                \STATE $\textbf{F}_u^g \gets M_u^g \odot \textbf{F}_u$
                \STATE Generate combination ratio $\gamma \in (0, 1)$\\
                \(\triangleright\) Total $L$ spatial locations in $\textbf{F}_i$\\
                \(\triangleright\) Draw with repeat and excluding all zeros feature
                \STATE $\{\textbf{f}_c^s\}$ $\gets$ Draw $\gamma L$ feature vectors from $\textbf{F}_c^s$
                \STATE $\{\textbf{f}_u^g\}$ $\gets$ Draw $(1 - \gamma) L$ feature vectors from $\textbf{F}_u^g$ 
                \STATE $\textbf{F}_c^{aug}$ $\gets$ Merge $\{\textbf{f}_c^s\}$ and $\{\textbf{f}_u^g\}$ 
                \STATE Append $\textbf{F}_c^{aug}$ to $b_{out}$
            \ENDFOR
            \STATE $\{\textbf{F}_k\}$ $\gets$ Draw $N_t(1 + N_a)$ samples from head classes
            \STATE Append $\{\textbf{F}_k\}$ to $b_{out}$
            
        \ENDFOR
        
	\end{algorithmic}
	\label{alg:feat_aug}
\end{algorithm}

The augmented samples are generated online to fine tune the network trained in Phase-I to improve the performance of the tail classes. In each batch, we sample $N_t$ images from the tail classes, and generate $N_a$ augmented samples online for each of the real samples, which creates a batch including $N_t(1 + N_a)$ samples from the tail classes. The same number of images are also randomly drawn from the head classes to balance the distribution. Thus, a batch of size $2N_t(1 + N_a)$ is generated online for each fine tuning iteration. We summarize this process in Alg.~\ref{alg:feat_aug}.

Fine tuning is performed on the layers after the features being extracted. Since the augmentation is conducted in the feature space, augmented samples can be generated at any stage of the network. However, the deeper features, compared to its shallow counterparts, are more linearly separable, which greatly help the fusion of features from the tail classes and their confusing classes. Moreover, richer spatial information in the lower-level feature maps may introduce artifacts to bias the model training. We analyze the detailed effect of augmenting samples at different depths in Sec.~\ref{sec:ablation}. We choose the features right before the last average pooling layer to help with the classification performance and at the same time to realize a simple design. Since the average pooling layer accumulates features in all spatial locations, the spatial distribution of class-generic and class-specific features become irrelevant in the augmented samples. Therefore, when combining, only the ratio between the two types of features needs to be given. Finally, we use the augmented batches to fine tune the FC classifier layer as shown in Fig.~\ref{fig:overview}.


\section{Experiments}
\label{exp}

We conduct experiments on the artificially created long-tailed CIFAR dataset~\cite{cui2019class} with various simulated imbalance factors, ImageNet-LT~\cite{liu2019large}, Places-LT~\cite{liu2019large} and the real world long-tailed iNaturalist 2017 and 2018~\cite{van2018inaturalist} datasets to validate the proposed method. Deep residual network (ResNet) with various depth are employed in our experiments.

\begin{figure}[t]
    \begin{minipage}{.47\linewidth}
	\centering
	\includegraphics[width=1\linewidth]{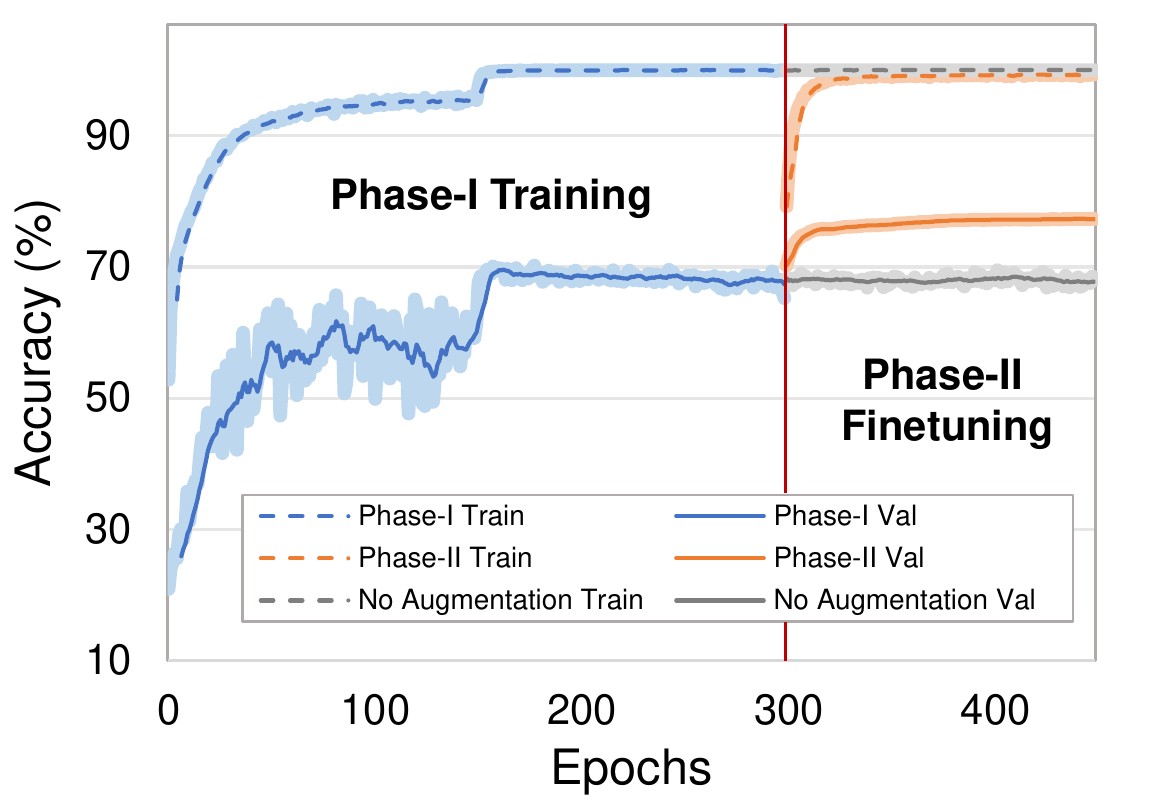}
	\caption{Learning Curve for long-tailed CIFAR-10 with an imbalance factor of 200 using ResNet-18. The overall accuracy of the validation dataset is illustrated.}
	\label{fig:learning-curve}
	\end{minipage}
    \hfill
    \begin{minipage}{.47\linewidth}
    \centering
	\begin{tabular}
	{c@{\hspace{0.3mm}}c@{\hspace{.4mm}}}
        \includegraphics[width=0.485\linewidth]{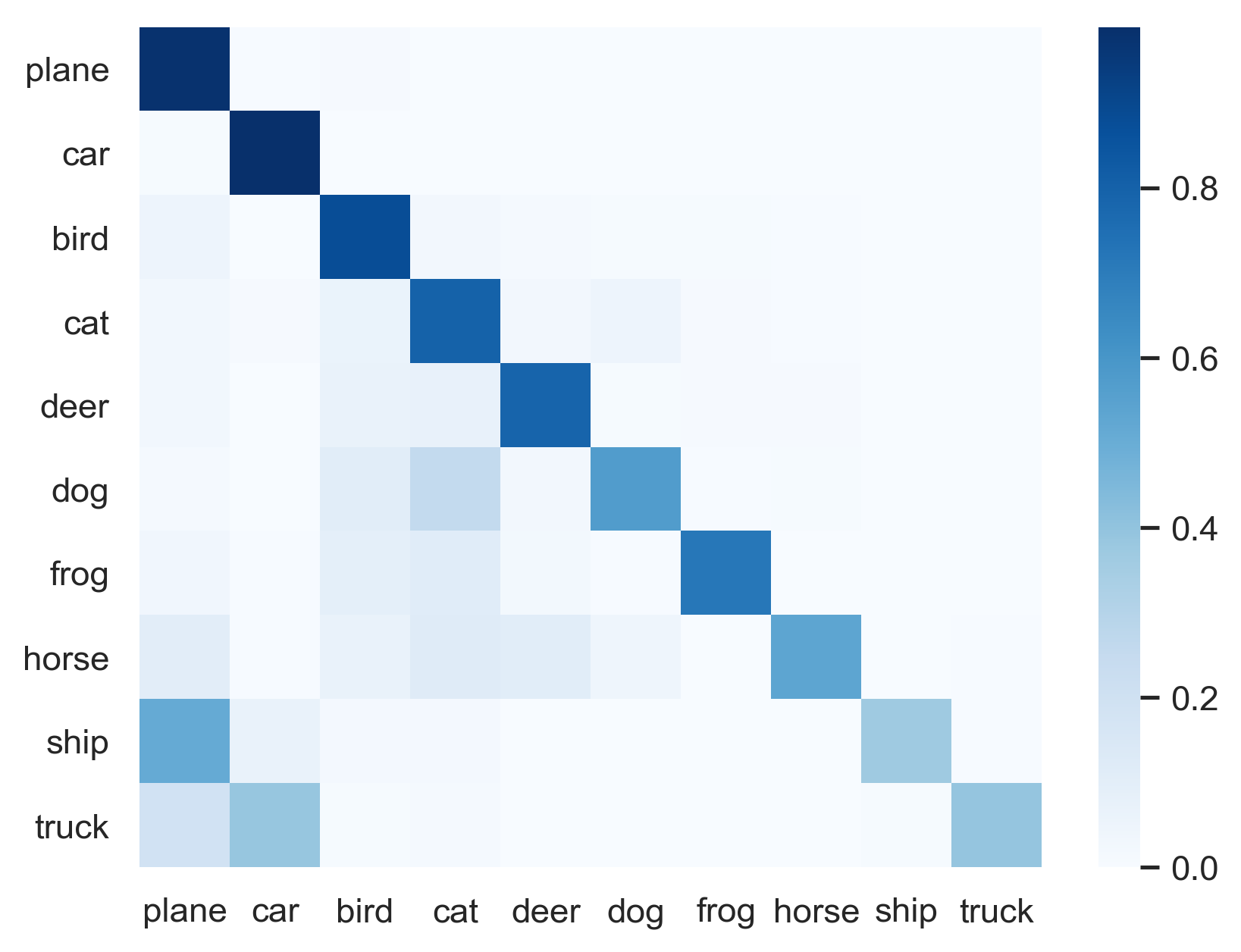} &
    	\includegraphics[width=0.5\linewidth]{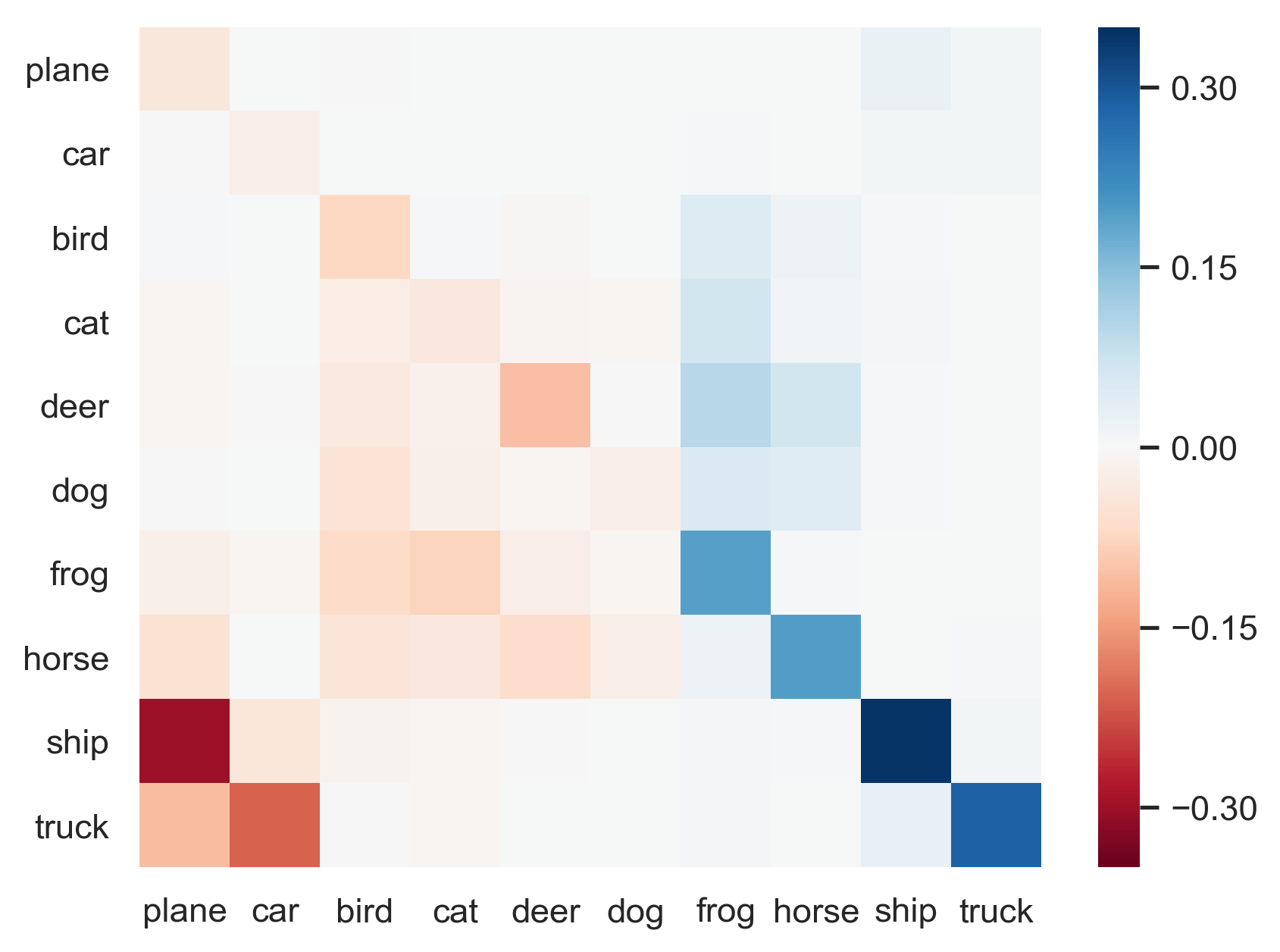}
    	\\
        \quad\includegraphics[width=0.45\linewidth]{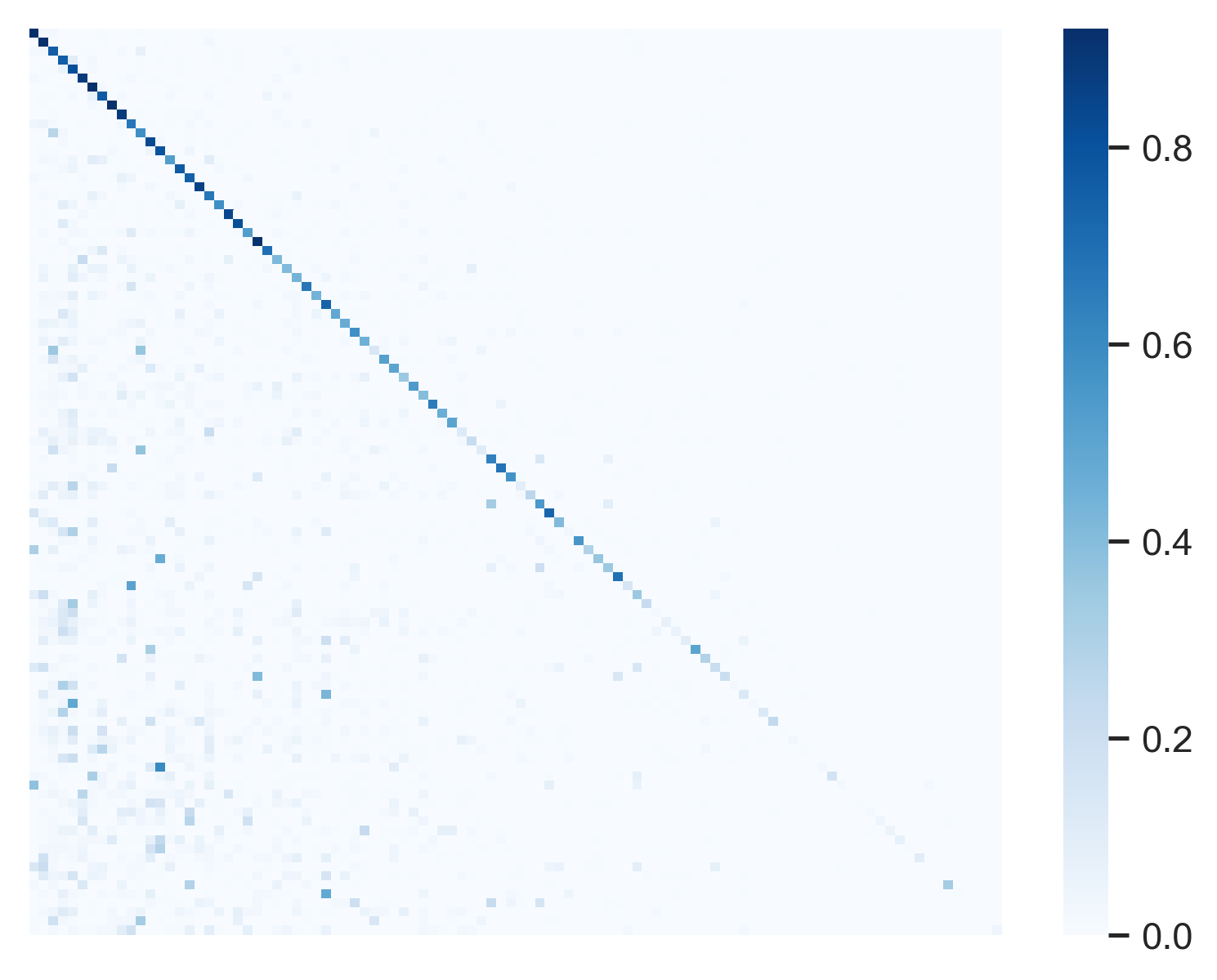} & \quad
    	\includegraphics[width=0.47\linewidth]{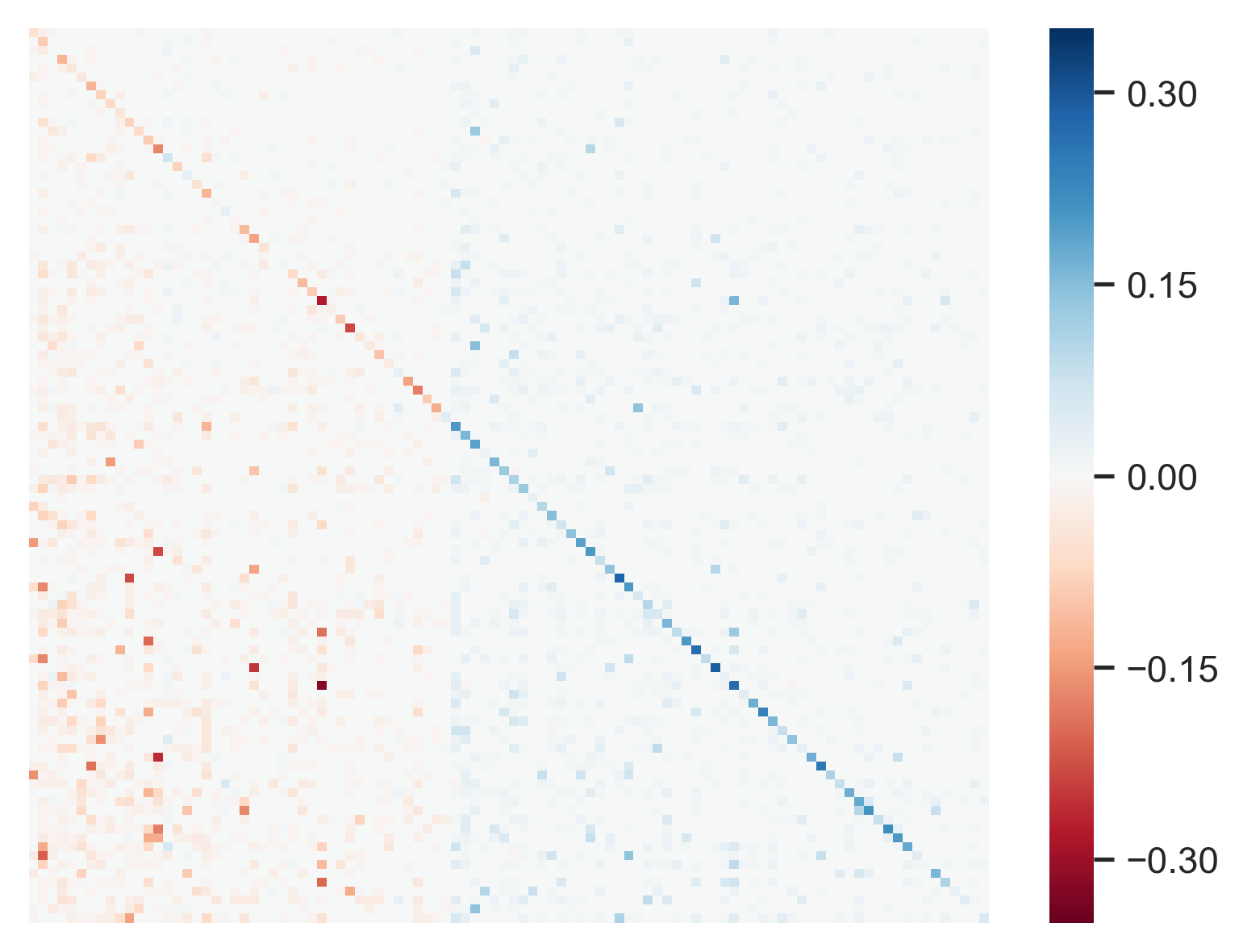}
    	\\
    	\scriptsize Phase-I & \scriptsize Changes After Phase-II
	\end{tabular}
    
    \caption{Confusion matrix for CIFAR-10 (upper) and CIFAR-100 (bottom) at IM 200 using ResNet-18.}%
    \label{fig:confusion_matrix}%
    \end{minipage}
\end{figure}

\subsection{Long-tailed CIFAR}

\begin{table*}[t]
    \centering
    \scriptsize
    \begin{tabular}{c|c|c|c|c|c|c|c|c|c|c}
        \hline\thickhline
         &  \multicolumn{5}{c|}{ResNet-18} & \multicolumn{5}{c}{ResNet-34}\\
         \hline
         IM & 10 & 20 & 50 & 100 & 200 & 10 & 20 & 50 & 100& 200\\
        \hline
        Baseline & 90.73 &87.24& 82.32& 75.16& 70.22 & 91.03 & 87.32& 82.74&78.58 & 71.42 \\
        \hline
        CB~\cite{cui2019class} $\beta=0.9$ & 90.79 & 86.61& 81.9& 75.16 & 69.16 & 91.03 & 87.18& 82.48& 75.99 &70.0\\
        \hline
        CB~\cite{cui2019class} $\beta=0.999$ & 90.54 & 86.83& 81.81& 76.4 & 69.83 & 90.74 & 87.24& 81.66 & 74.85 & 70.08\\
        \hline
        CB~\cite{cui2019class} $\beta=0.9999$ & 89.61 & 86.05& 80.4& 75.04 & 69.21 & 90.69 & 86.9& 81.06& 75.74 & 68.79\\
        \hline
        FL~\cite{lin2017focal} $\gamma=0.5$ & 90.66 &86.61& 81.55& 74.99 & 69.06 & 90.76 &87.18& 81.91& 76.5 &69.87\\
        \hline
        FL~\cite{lin2017focal} $\gamma=1.0$ & 90.59 &86.83& 81.79& 74.07 & 68.23 & 90.7 &87.24& 81.34& 76.44 & 70.02\\
        \hline
        FL~\cite{lin2017focal} $\gamma=2.0$ & 90.5 &86.05& 81.25& 75.13 & 68.27 & 90.08 &86.9& 82.44& 75.58 & 69.87\\
        \hline
        SLA~\cite{lee2019rethinking} & - & - & - & - & - & 89.58 & - & - & 80.24 & -\\
        \hline
        Ours & \textbf{91.75} & \textbf{88.54}& \textbf{84.51}& \textbf{80.57} & \textbf{77.06} & \textbf{91.2} & \textbf{89.26}& \textbf{84.49}& \textbf{82.06} & \textbf{75.52} \\  
        \hline\thickhline
    \end{tabular}
    \caption{Classification accuracy on long-tailed CIFAR-10.}
    \label{tab:cifar10}
    
    \begin{tabular}{c|c|c|c|c|c|c|c|c|c|c}
        \hline\thickhline
         &  \multicolumn{5}{c|}{ResNet-18} & \multicolumn{5}{c}{ResNet-34}\\
         \hline
        IM & 10 & 20 & 50 & 100 & 200 & 10 & 20 & 50 & 100& 200\\
        \hline
        Baseline & 62.59 & 57.09& 48.55& 43.65& 38.87 & 63.87 & 57.55& 48.07& 43.55 & 37.5 \\
        \hline
        CB~\cite{cui2019class} $\beta=0.9$ & 63.1 & 57.02& 48.15&43.51 & 38.58 & 64.14 & 58.03& 48.44&42.94 & 38.84\\
        \hline
        CB~\cite{cui2019class} $\beta=0.999$& 61.76 & 55.3& 44.28&32.19 & 26.61 & 63.05 & 54.13& 40.89&32.65 & 26.2\\
        \hline
        CB~\cite{cui2019class} $\beta=0.9999$& 60.71 & 53.93& 42.02&31.32 & 25.91 & 62.28 & 53.64& 40.03&29.82 & 26.63\\
        \hline
        FL~\cite{lin2017focal} $\gamma=0.5$ & 62.64 & 57.02& 47.9&42.82 & 38.73 & 64.36 &58.45 & 48.31&42.72 & 36.18\\
        \hline
        FL~\cite{lin2017focal} $\gamma=1.0$& 62.85 & 57.22& 47.76&42.81 & 40.47 & 64.83 & 58.78& 48.24&42.64 & 37.29\\
        \hline
        FL~\cite{lin2017focal} $\gamma=2.0$& 63.37 &57.15&  47.0&42.18 & 40.31 & 64.48 & 58.55& 47.47&43.33 & 38.11\\
        \hline
        SLA~\cite{lee2019rethinking} & - & - & - & - & - & 59.89 & - & - & 45.53 & -\\
        \hline
        Ours & \textbf{65.08} & \textbf{58.69}& \textbf{51.9} &\textbf{46.57} & \textbf{42.84} & \textbf{65.29} & \textbf{59.75}& \textbf{52.17}& \textbf{48.51} & \textbf{41.46} \\  
        \hline\thickhline
    \end{tabular}
    \caption{Classification accuracy on long-tailed CIFAR-100.}
    \label{tab:cifar100}
\end{table*}

To demonstrate the effectiveness of the proposed method, long-tailed versions of CIFAR dataset are generated following the protocol mentioned in~\cite{cui2019class} as $IM = \max(\{N_i\})/\min(\{N_i\})$. Five datasets of different imbalance factors, \{10, 20, 50, 100, 200\}, are created for both CIFAR-10 and CIFAR-100, where an imbalance factor is defined as 
where $N_i$ is the number of training samples of the $i$-th class. ResNet with depth 18 and 34 are adapted for this experiment. We use the original validation set of the CIFAR-10 and CIFAR-100 to evaluate the performance. 

The baseline network and the proposed method are implemented in PyTorch and run on a Xeon CPU of 2.1 GHz and Tesla V100 GPU. The initial learning rate for Phase-I is 0.1 and decreases by $1/10$ every 150 epochs. The feature sub-network and base classifier are trained for 300 epochs. In Phase-II, the learning rate is fixed at 0.001. The classifier is fine tuned for 6,400 iterations with a batch size of 128. For each real sample in tail classes, we choose $N_a = N_f = 3$.

A sample learning curve for long-tailed CIFAR-10 with an imbalance factor of 200 using ResNet-18 is shown in Fig.~\ref{fig:learning-curve}. The performance of the proposed method is compared with a baseline setting of the same learning rate but without the feature space augmentation. After the Phase-II feature space augmentation, the accuracy of the proposed method on the validation set increases about $7\%$ during the fine tuning stage, while no noticeable change in accuracy for the baseline setting is observed.

To further illustrate the improvement, the confusion matrix before Phase-II and its changes are shown in Fig.~\ref{fig:confusion_matrix}. After Phase-I training, tail classes show poor accuracy on the validation set due to insufficient training samples in those classes. Most mis-classified samples fall into the first several head classes, where most training samples belong to, as indicated in the left bottom corner of the confusion matrix. After Phase-II fine tuning, significant improvement is observed for the diagonal elements of the tail classes. The off-diagonal elements decrease accordingly. Although the accuracy of the head classes decrease slightly, due to dramatic improvement in the tail classes, the overall accuracy still increases.

The complete classification performance on different imbalance factors of the two dataset are shown in Tab.~\ref{tab:cifar10} and \ref{tab:cifar100}. The method using the same ResNet with cross-entropy loss and traditional data augmentation on input images is referred as Baseline in Tab.~\ref{tab:cifar10} and \ref{tab:cifar100}. In our experiments, we compare our method with the state of the arts on addressing the long-tailed problem, including Class-balanced (CB) loss~\cite{cui2019class} based method and Focal Loss (FL) from~\cite{lin2017focal} with various choices of hyper-parameters and augmentation based method~\cite{lee2019rethinking}. Our method outperforms all other methods in both datasets.

\subsection{ImageNet-LT and Places-LT Dataset}

We also evaluate the proposed method on two constructed large-scale long-tailed datasets ImageNet-LT and Places-LT~\cite{liu2019large}. ImageNet-LT is a subset of the ILSVRC2012 dataset. Its training set is drawn from the original training set following the Pareto distribution with $\alpha=6$, which results  115.8K images from 1000 categories with a maximum of 1280 images per category and a minimum of 5 images per category ($IM=256$). The original validation set with balanced 50K images is used as test set in our experiments. Places-LT dataset is constructed similarly with ImageNet-LT from Places-2 dataset. Finally, 184.5K images from 365 categories are collected, where the largest class contains 4980 images while the smallest ones with 5 images ($IM=996$). The test set contains balanced 36.5K images.

For fair comparison, we use the same scratch ResNet-10 for ImageNet-LT and pre-trained ResNet-152 for Places-LT as in \cite{liu2019large}. The numerical results and comparison with other peer methods are reported in Tab.~\ref{tab:img}. We also evaluate the combination of other balanced loss methods with the proposed method in these experiments. The different losses are applied in the Phase-I training. We use ``Ours+FL'' to refer the experiments using Focal Loss and ``Ours'' for ordinary cross-entropy loss. Both of our methods achieve comparable performance with the state-of-the-art method. 

Note that, ``Ours+FL" shows better performance than ``Ours'' in the ImageNet-LT dataset while ``Ours'' is better in Places-LT. Feature maps generated in the shallow network as ResNet-10 is not as sparse as in ResNet-152. Therefore, as explained in Sec.~\ref{sec:online_aug}, the feature space augmentation delivers more performance boost to ResNet-152. On the other hand, our class balanced training batch generation achieves a similar effect as other balanced loss methods in the Phase-II fine tuning. But applying those losses in the Phase-I may still improve the performance when poor Phase-I performance affects CAM quality.

\begin{table}[t]
\scriptsize
\centering
\begin{tabular}{l|c|c|c|c|c|c|c|c}
\hline\thickhline
& \multicolumn{4}{c}{\textbf{ImageNet-LT}} & \multicolumn{4}{|c}{\textbf{Places-LT}} \\ \hline
 &  \begin{tabular}{@{}c@{}}$>100$ \\ Many\end{tabular}   &  \begin{tabular}{@{}c@{}}$ \leqslant 100 \; \& >20$ \\ Medium\end{tabular}   &  \begin{tabular}{@{}c@{}}$<20$ \\ Few\end{tabular}   &  Overall &  \begin{tabular}{@{}c@{}}$>100$ \\ Many\end{tabular}   &  \begin{tabular}{@{}c@{}}$ \leqslant 100 \; \& >20$ \\ Medium\end{tabular}   &  \begin{tabular}{@{}c@{}}$<20$ \\ Few\end{tabular}   &  Overall  \\ \hline
 Plain Model~\cite{liu2019large}&   40.9  &  10.7   &  0.4   &       20.9  &   \textbf{45.9}  &  22.4   &  0.36   &       27.2  \\ \hline
 Lifted Loss~\cite{oh2016deep}&  35.8   &   30.4   &   17.9  &      30.8  &  41.1   &    35.4   &    24  &       35.2   \\ \hline
 FL~\cite{lin2017focal}&  36.4   &   29.9  &   16   &      30.5   &  41.1   &   34.8  &  22.4   &      34.6  \\ \hline
 Range Loss~\cite{zhang2017range}&  35.8   &   30.3  &  17.6   &     30.7  &  41.1   &    35.4  &  23.2   &     35.1   \\ \hline
 FSLwF~\cite{gidaris2018dynamic}&  40.9   &    22.1  &   15   &     28.4   &  43.9  &    29.9  &   \textbf{29.5}   &     34.9    \\ \hline
 OLTR ~\cite{liu2019large} &   43.2  &  \textbf{35.1}   &  \textbf{18.5}   &     \textbf{35.6}   &   44.7   &  37  &  25.3   &     35.9     \\ \hline
 Ours &   \textbf{47.3}  &   31.6  &   14.7  &     35.2   &   42.8  &   \textbf{37.5}  &   22.7  &     \textbf{36.4}    \\ \hline
 Ours+FL &   47.0  &   31.3  &   16.8  &     35.3   &   42.2 &   36.4  &   24.0  &     36.0   \\ \hline
 \thickhline
\end{tabular}
\caption{Top-1 classification accuracy on ImageNet-LT and Places-LT.}
\label{tab:img}

\label{tab:places}
\end{table}

\subsection{iNaturalist}

iNaturalist is a real-world fine-grained species classification dataset. Its 2017 version contains 579,184 training images of 5,089 categories, and its 2018 version~\cite{van2018inaturalist} has 437,513 training samples in 8,142 classes. The imbalance factor for iNaturalist 2017 is 435 and 500 for iNaturalist 2018. For both versions, there are three validation samples for each class. We adapt ResNet-50, ResNet-101 and ResNet-152 in our experiments, all with $224 \times 224$ input image size. The similar training strategy with CIFAR datasets is adapted for iNaturalist. In Phase-I, the starting learning rate is 0.1 and reduced every 30 epochs for total of 100 epochs. In Phase-II, fine tuning is performed with a fixed learning rate of 0.001 for 200 iterations. The top-1 classification accuracy for the validation set of the two datasets are reported in Tab.~\ref{tab:inat}. We compare the proposed method with class-balanced cross-entropy loss on ResNet-152 and class-balanced focal loss on ResNet-101/50. Our method has shown the best performance in all the settings.

\begin{table}[t]
    \centering
    \scriptsize
    \begin{tabular}{c|c|c|c}
        \hline\thickhline
         &  & iNaturalist 2017 & iNaturalist 2018\\
         \hline
        \multirow{3}{*}{Baseline} & ResNet-50 & 60.50  & 62.27  \\
        \cline{2-4}
        & ResNet-101 & 61.81  & 65.19  \\
        \cline{2-4}
        & ResNet-152 & 65.12 & 66.17 \\
        \hline
        \multirow{3}{*}{CB~\cite{cui2019class}} & ResNet-50$^*$ &58.08 & 61.12\\
        \cline{2-4}
        & ResNet-101$^*$ & 60.94 & 63.88 \\
        \cline{2-4}
        & ResNet-152 & 64.75 & 66.97 \\
        \hline
        \multirow{3}{*}{Ours} & ResNet-50 &\textbf{61.96} &\textbf{65.91} \\
        \cline{2-4}
        & ResNet-101 & \textbf{64.16}  & \textbf{68.39}  \\
        \cline{2-4}
        & ResNet-152 &\textbf{66.58} &\textbf{69.08} \\
        
        \hline\thickhline
    \end{tabular}

    \caption{Top-1 classification accuracy on iNaturalist ($^*$: results from literature).}
    \label{tab:inat}
\end{table}

\subsection{Ablation Analysis}
\label{sec:ablation}

\begin{figure}[t]
    \begin{minipage}{.485\linewidth}
	\centering
	\includegraphics[width=1\linewidth]{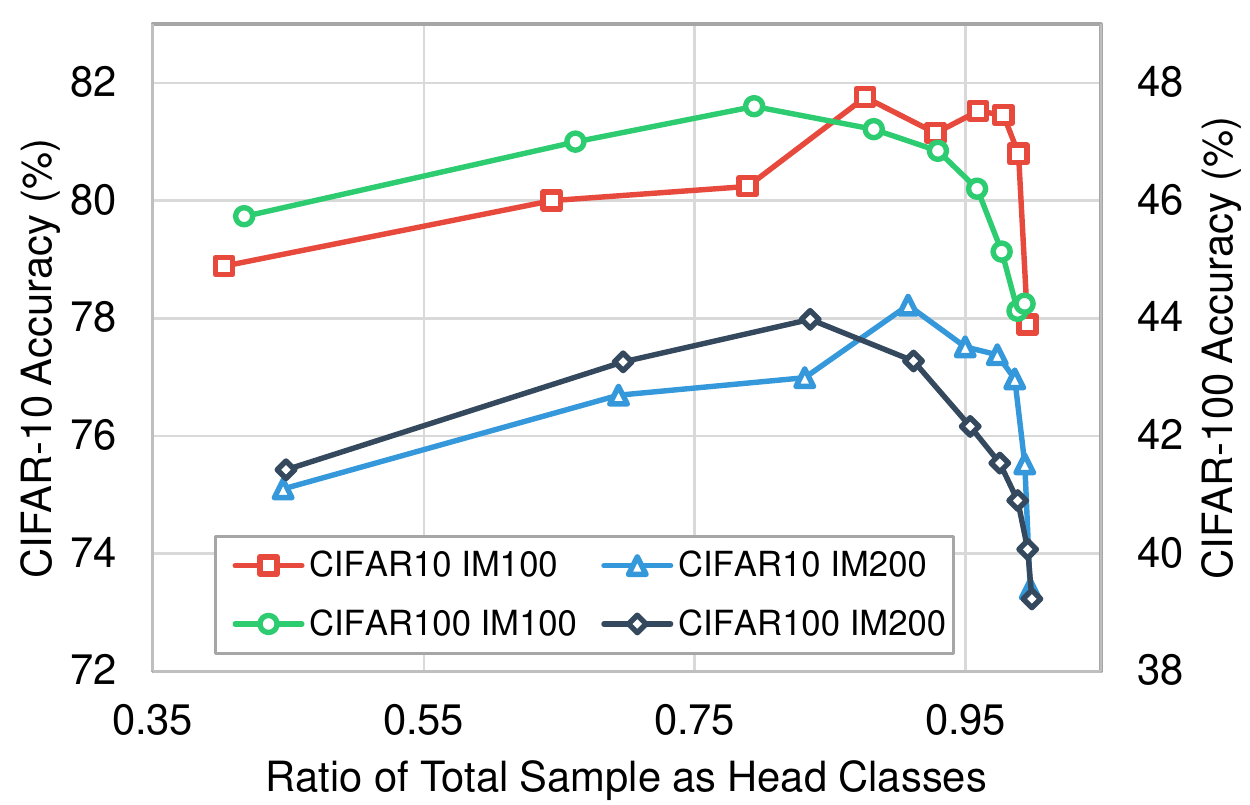}
	\caption{Phase-II performance dependence on the ratio of total training samples used as the head class sample.}
	\label{fig:head_class}
	\end{minipage}
	\hfill
    \begin{minipage}{.48\linewidth}
	\centering
	\includegraphics[width=1\linewidth]{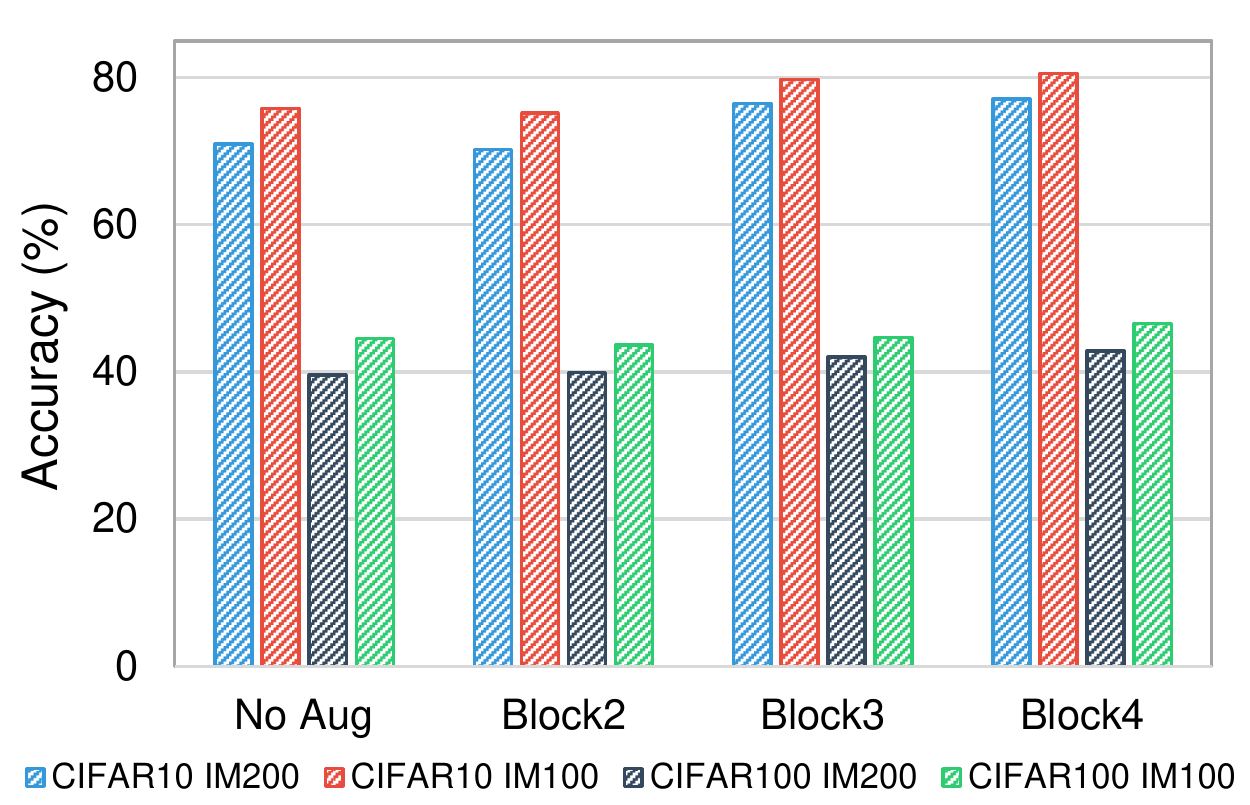}
	\caption{Classification accuracy by applying feature space augmentation at different depth of ResNet-18.}
	\label{fig:location}
	\end{minipage}
\end{figure}

One major hyper-parameter in the proposed method is how to separate the head classes from the tail classes. Specifically, the classes are first sorted in the descent order by the number of training samples in each class as illustrated in Fig.~\ref{fig: longtail}. The first $h$ classes are chosen as head classes. In order to unify the choice between different imbalance factors of datasets, we introduce $h_r \in (0, 1) $ which is the ratio between the number of samples in the head classes and the total number of samples. Different $h_r$ choices against the Phase-II classification accuracy are evaluated in Fig.~\ref{fig:head_class}. The curves among different datasets show peaks around $h_r = 0.95$. On the left of peaks, fewer samples or classes are used as the head class, and thus class-generic features cannot be drawn sufficiently for feature augmentation. On the right side, fewer classes are selected as the tail classes, and therefore some classes with insufficient training samples will not be fine tuned with augmented samples. For consistency, we choose the minimum of $h$ that satisfies $h_r \ge 0.9$ in all the CIFAR experiments.

We also investigate the classification performance when applying the feature space augmentation at different depths of the network. 
The ResNet architecture we adapted usually consists of four convolutional blocks. We plug our feature space augmentation after each of the last three convolutional blocks of 
ResNet-18 on CIFAR dataset. When augmenting features after Block2 and Block3, class-specific features in $\textbf{F}_u^g$ are replaced with the class-specific features in $\textbf{F}_c^g$ with random ratio to generate augmented samples, where spatial information of $\textbf{F}_u^g$ is preserved. The corresponding classification accuracy after Phase-II fine tuning is shown in Fig.~\ref{fig:location}. From Fig.~\ref{fig:location}, one can observe that feature augmentation after Block4 gains the best performance among different datasets and imbalance factors. The feature maps closer to the input side contain more spatial information, 
which also introduces additional artifacts into the augmented samples. Features generated by Block4 are directly passed into the global pooling layer where the noise in the spatial dimension introduced by augmentation can be eliminated. Moreover, the linearity of the high-level feature space helps the final linear operation of the fusion. We, therefore, apply the feature space augmentation after Block4. We also compare the performance of only sampling balanced finetuning batch without augmentation applied, which is refered as ``No Aug'' in Fig.~\ref{fig:location}.

\section{Conclusion}

In this paper, we propose a novel learning scheme to address the problem of training the deep convolutional neural network based classifier with long-tailed datasets. In detail, by combining the class-generic features in head classes with class-specific features in tail classes, augmented samples are online generated to enhance the performance of the tail classes. Results on long-tailed version CIFAR-10/100, ImageNet-LT, Places-LT and real-world long-tailed iNaturalist 2017/2018 datasets have shown the effectiveness of proposed method. 


\clearpage
%
%
\bibliographystyle{splncs04}
\bibliography{egbib}

\newpage

\section{Supplementary}

In the supplementary material, we first present the learning curve of different network architectures on CIFAR-10 and CIFAR-100. We have observed a significant improvement compared to the baseline model training using conventional data augmentation (used in ImageNet ResNet model training) across different network architectures and dataset. 
\begin{figure}[ht]
\begin{subfigure}{.5\textwidth}
  \centering
  \includegraphics[width=1\linewidth]{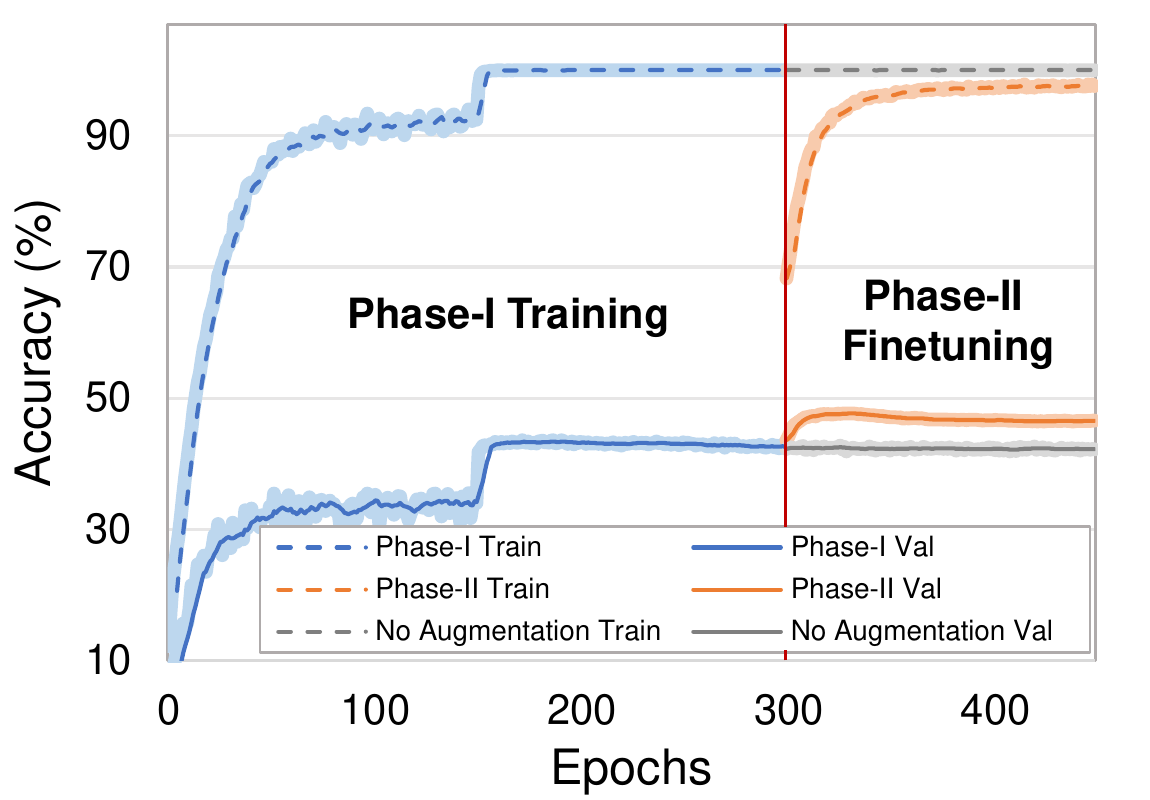}  
  \caption{CIFAR-100 with IM factor of 100 using ResNet-18}
  \label{fig:sub-first}
\end{subfigure}
\begin{subfigure}{.5\textwidth}
  \centering
  \includegraphics[width=1\linewidth]{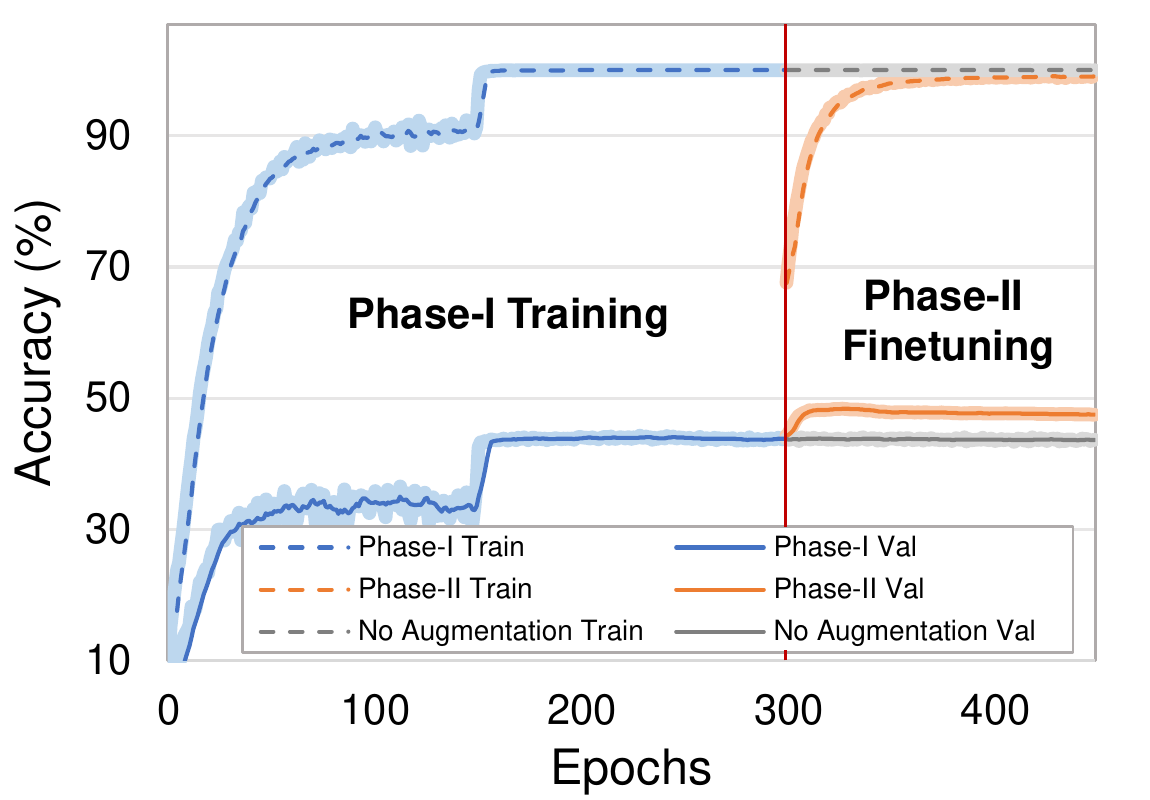}  
  \caption{CIFAR-100 with IM factor of 100 using ResNet-34}
  \label{fig:sub-second}
\end{subfigure}
\newline
\begin{subfigure}{.5\textwidth}
  \centering
  \includegraphics[width=1\linewidth]{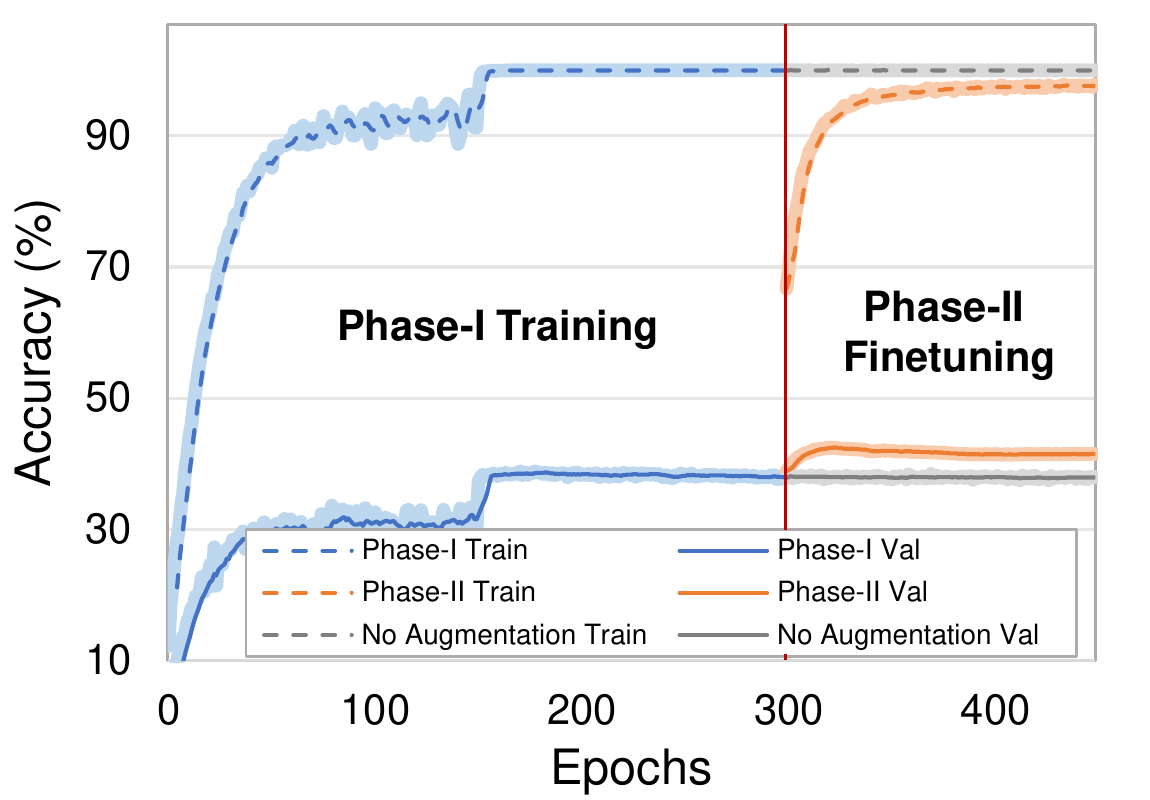}  
  \caption{CIFAR-100 with IM factor of 200 using ResNet-18}
  \label{fig:sub-second}
\end{subfigure}
\begin{subfigure}{.5\textwidth}
  \centering
  \includegraphics[width=1\linewidth]{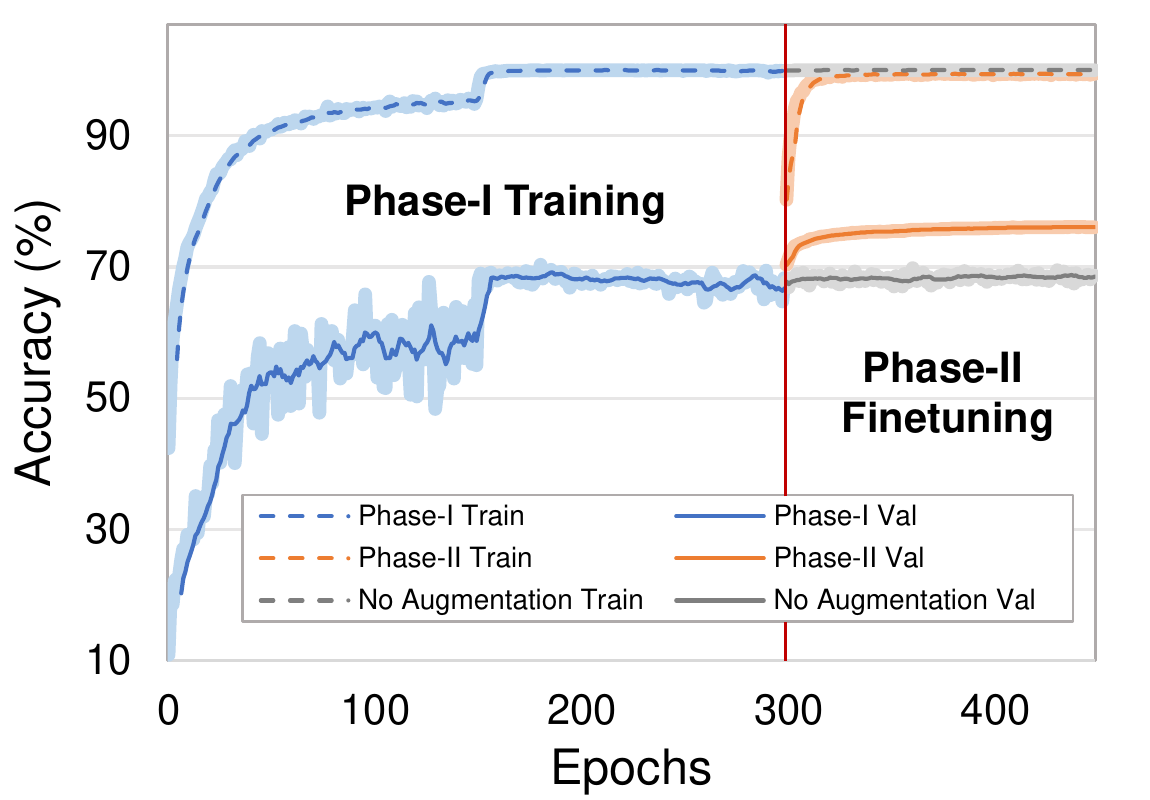}  
  \caption{CIFAR-10 with IM factor of 200 using ResNet-34}
  \label{fig:sub-second}
\end{subfigure}
\caption{Learning Curve}
\label{fig:fig}
\end{figure}

\subsection{Feature Space Visualization}
As discussed in Section 3.2,  We present the scatter plot of feature decomposition of different network architectures on CIFAR-10 and CIFAR-100. It shows a generic trend that the class-specific features are significantly more separated compared to class-generic features from the same class. This result further substantiate the assumption that after separating the class-specific features, the remaining class-generic features from the head classes can be helpful to recover the loss information of the tail classes.

\begin{figure}%
    \centering
    \vspace{-3px}
    \begin{subfigure}{.49\linewidth}
	\begin{tabular}
	{c@{\hspace{0.1mm}}c@{\hspace{.1mm}}}
        \includegraphics[width=0.5\linewidth]{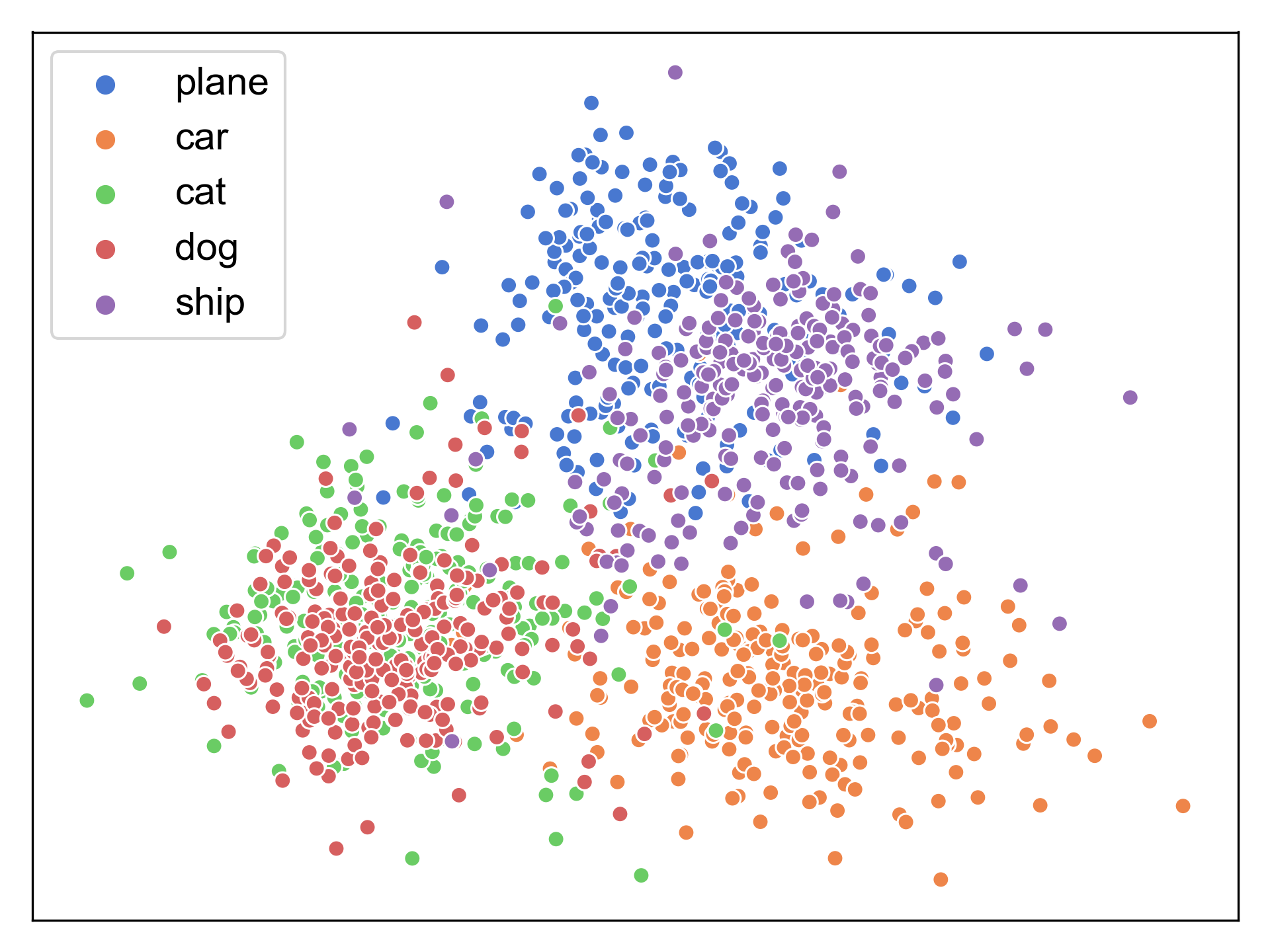} &
    	\includegraphics[width=0.5\linewidth]{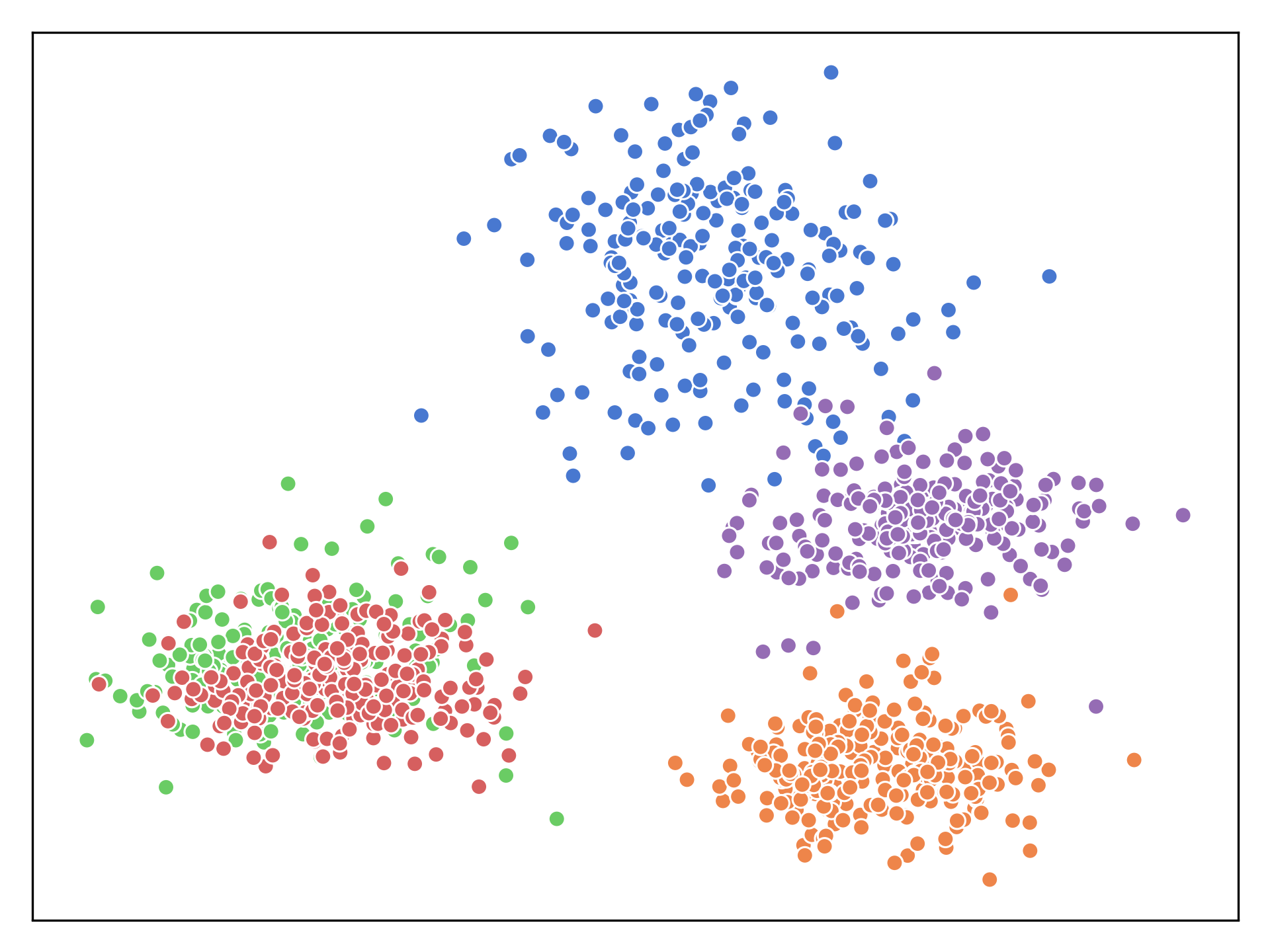}
    	\\
    	\small Class-Generic & \small Class-Specific
	\end{tabular}
	\caption{ResNet-34}
    \end{subfigure}
    \hspace{0.1mm}
    \begin{subfigure}{.49\linewidth}
	\begin{tabular}
	{c@{\hspace{0.1mm}}c@{\hspace{.1mm}}}
        \includegraphics[width=0.5\linewidth]{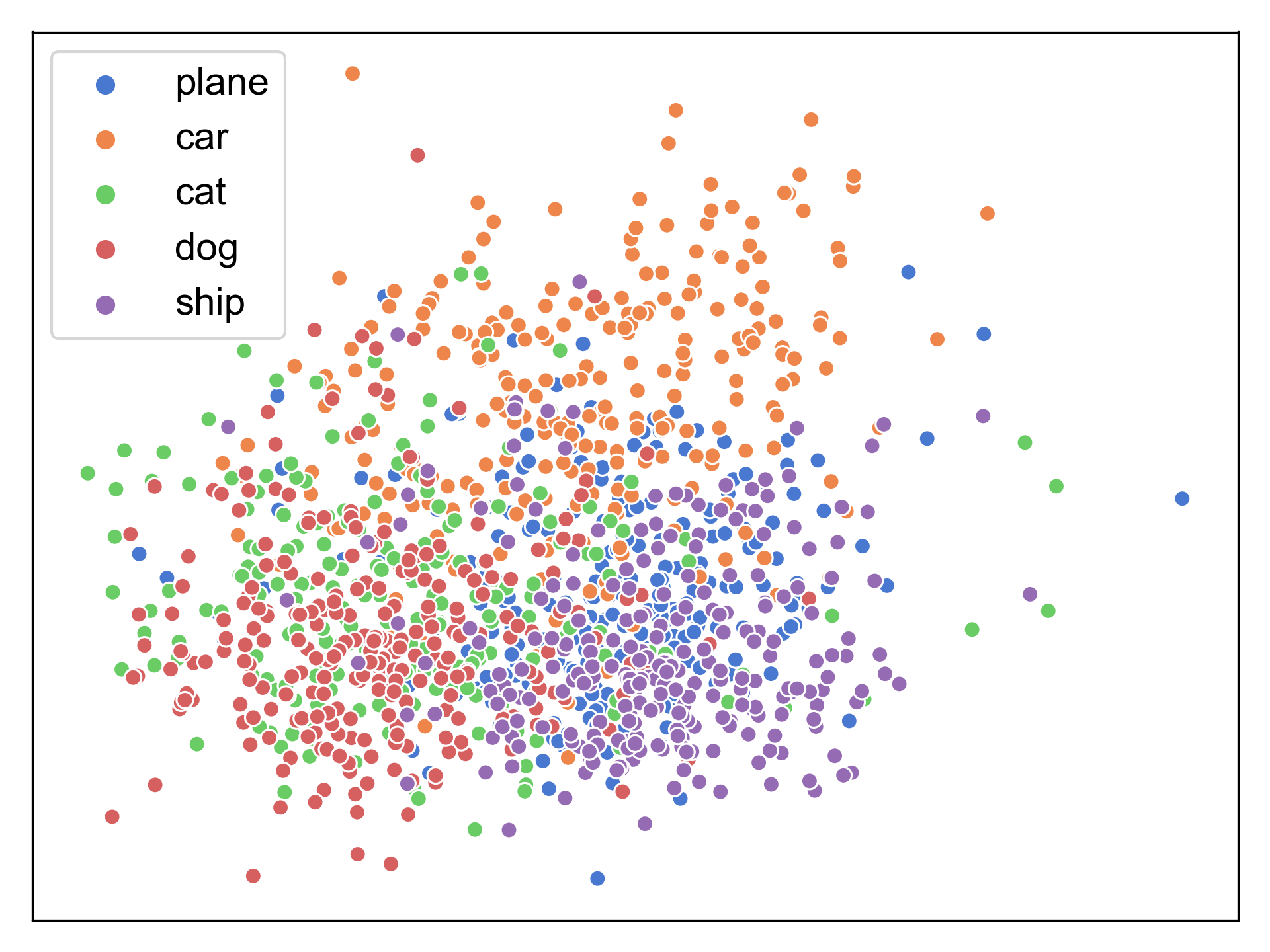} &
    	\includegraphics[width=0.5\linewidth]{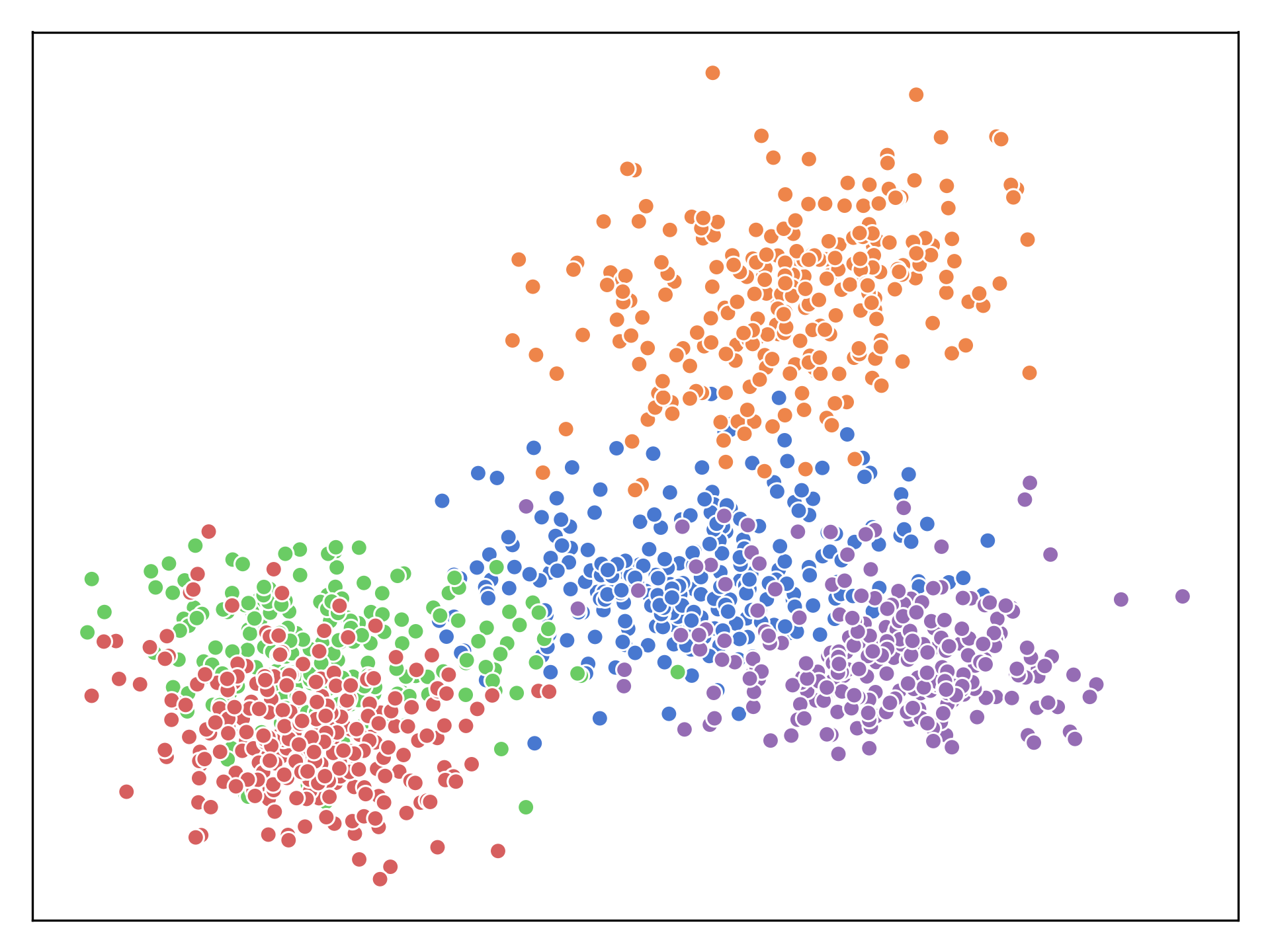}
    	\\
    	\small Class-Generic & \small Class-Specific
	\end{tabular}
	\caption{DenseNet-121}
    \end{subfigure}
    \newline
        \begin{subfigure}{.49\linewidth}
	\begin{tabular}
	{c@{\hspace{0.1mm}}c@{\hspace{.1mm}}}
        \includegraphics[width=0.5\linewidth]{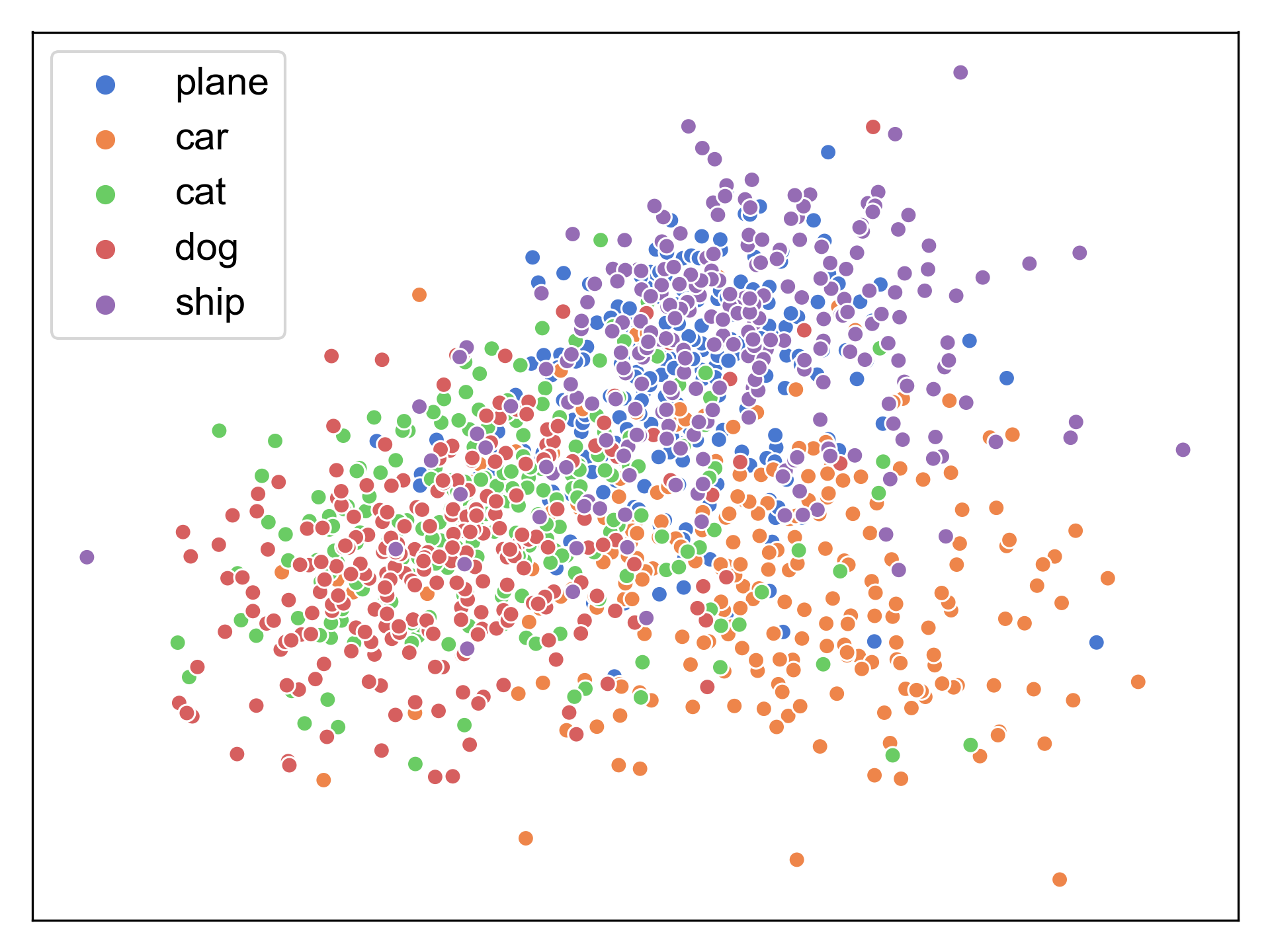} &
    	\includegraphics[width=0.5\linewidth]{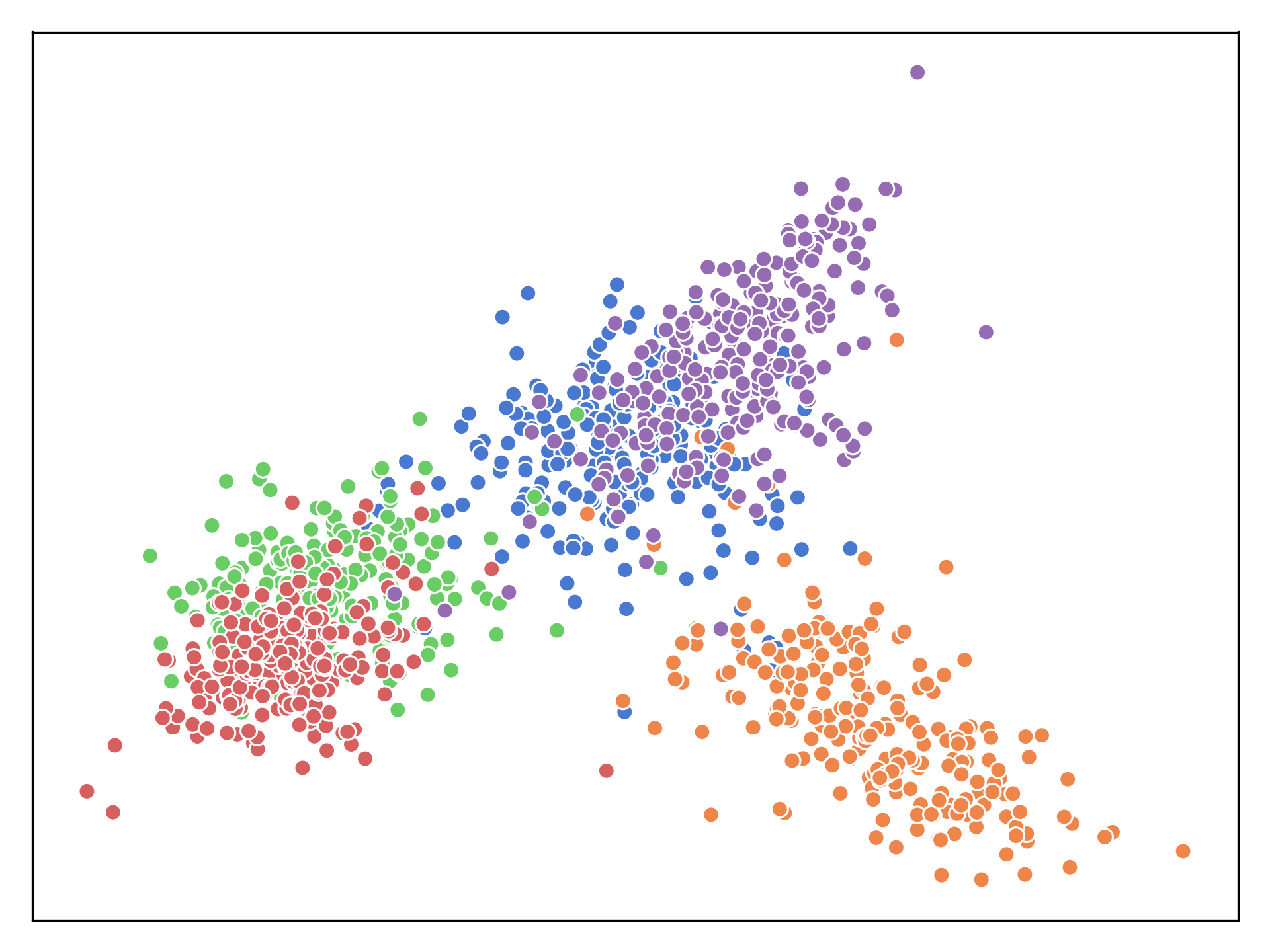}
    	\\
    	\small Class-Generic & \small Class-Specific
	\end{tabular}
	\caption{MobileNetv2}
    \end{subfigure}
    \hspace{0.1mm}
    \begin{subfigure}{.49\linewidth}
	\begin{tabular}
	{c@{\hspace{0.1mm}}c@{\hspace{.1mm}}}
        \includegraphics[width=0.5\linewidth]{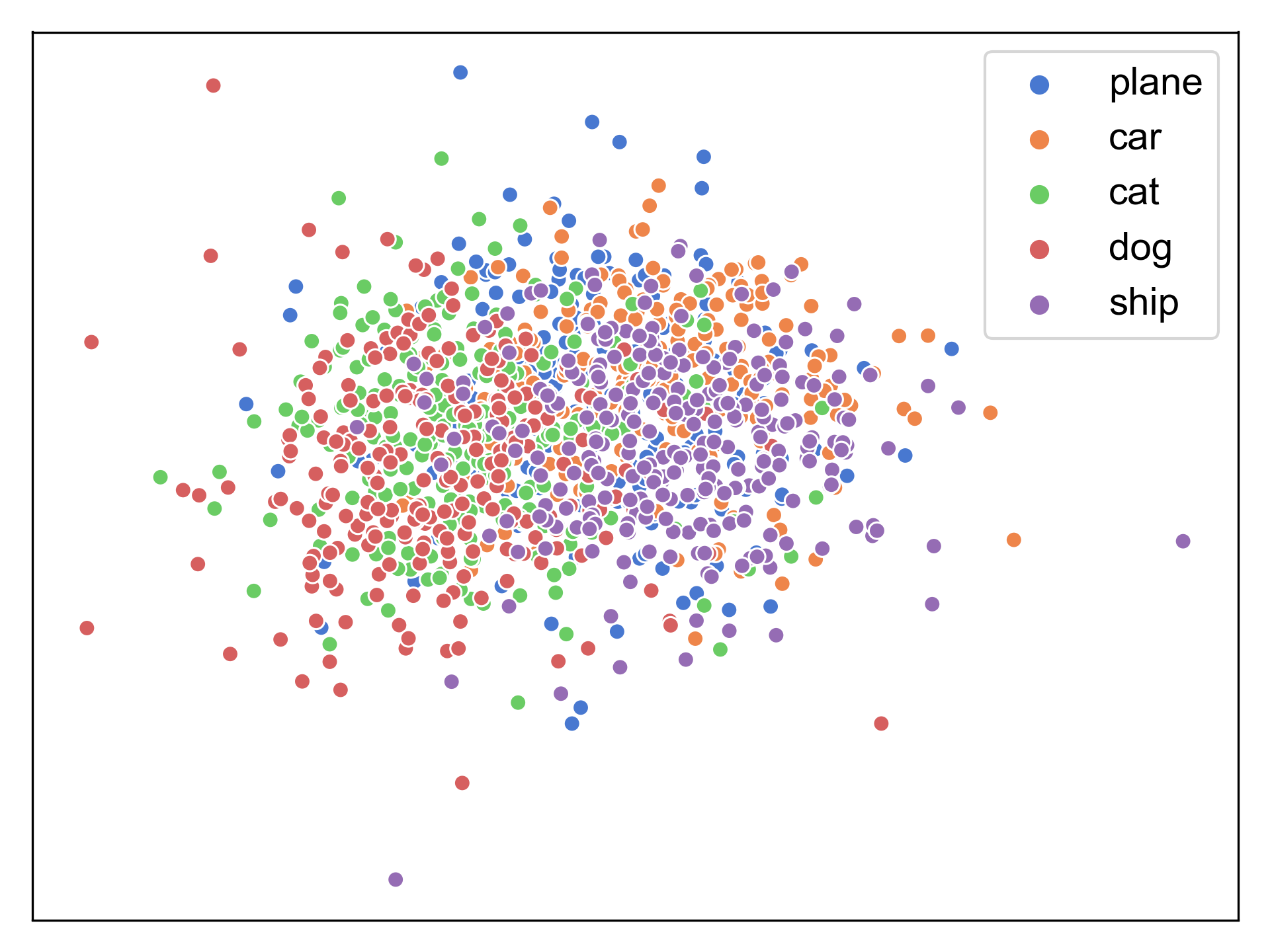} &
    	\includegraphics[width=0.5\linewidth]{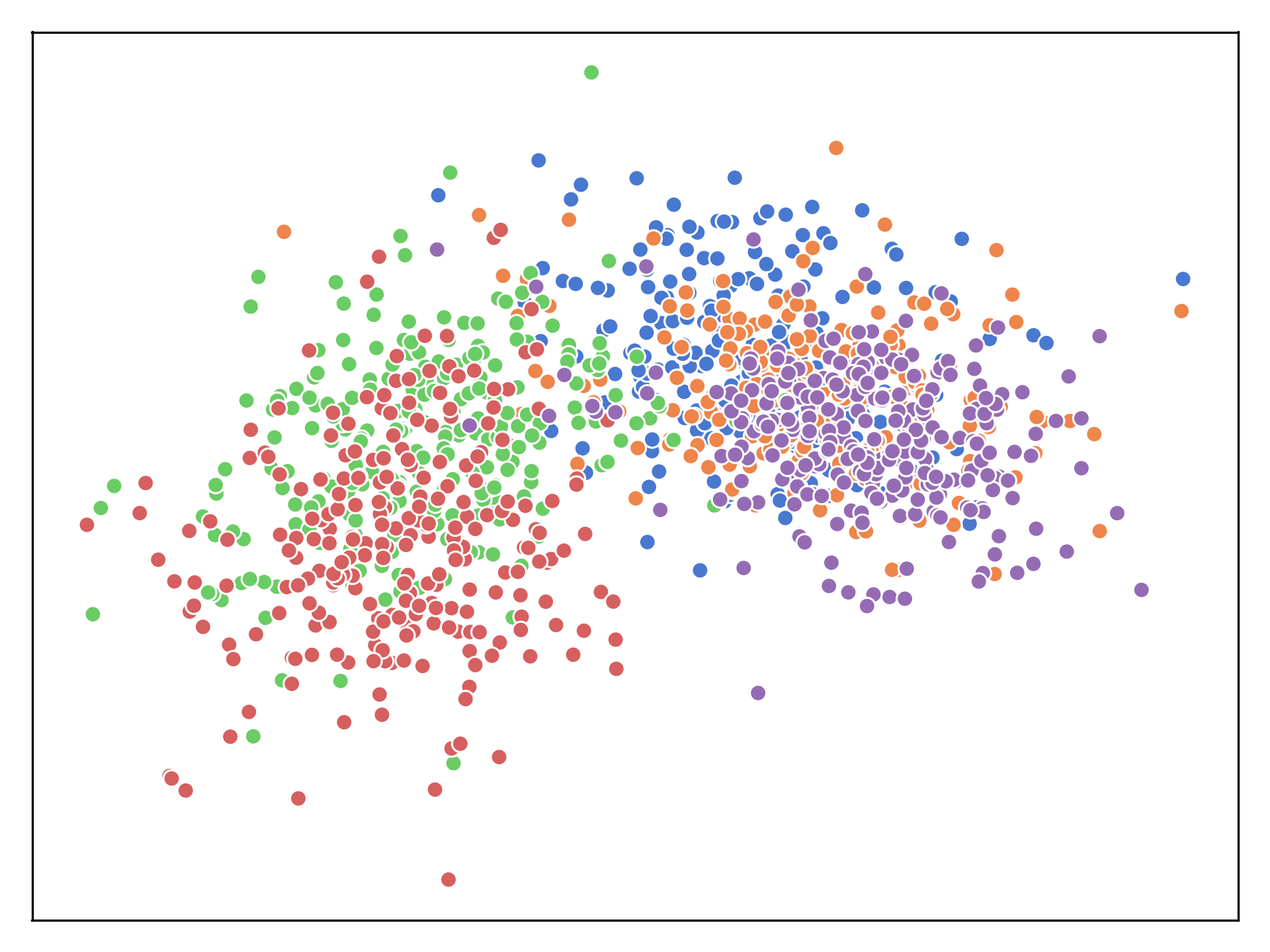}
    	\\
    	\small Class-Generic & \small Class-Specific
	\end{tabular}
	\caption{EfficientNet-B0}
    \end{subfigure}
    \vspace{-5px}
    \caption{CIFAR-10 Feature space visualization for different network architectures. }%
    \label{fig:confusion_matrix}%
    \vspace{-12px}
\end{figure}
\begin{figure}%
    \centering
    \vspace{-12px}
    \begin{subfigure}{.49\linewidth}
	\begin{tabular}
	{c@{\hspace{0.1mm}}c@{\hspace{.1mm}}}
        \includegraphics[width=0.5\linewidth]{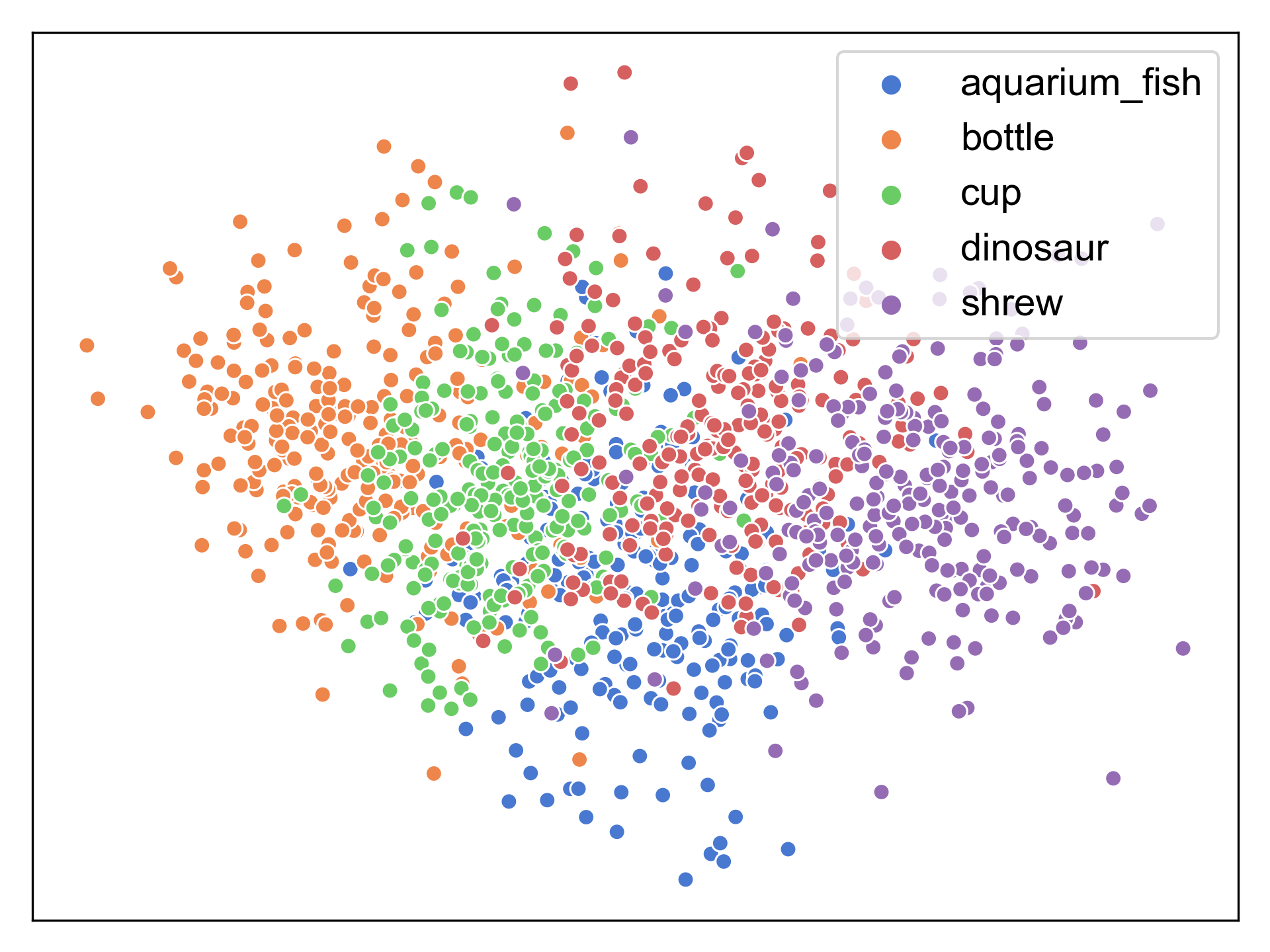} &
    	\includegraphics[width=0.5\linewidth]{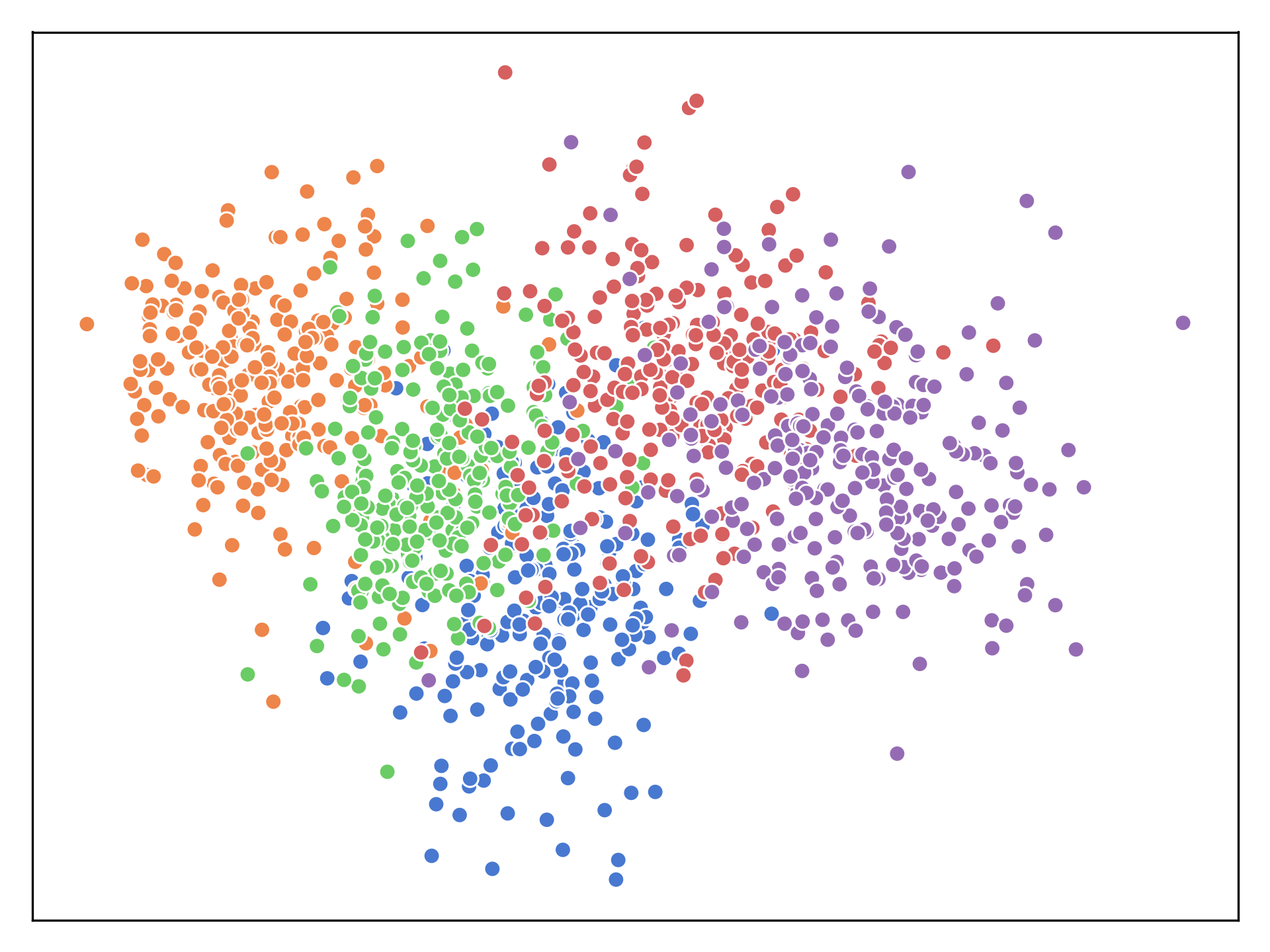}
    	\\
    	\small Class-Generic & \small Class-Specific
	\end{tabular}
    \end{subfigure}
    \hspace{0.1mm}
    \begin{subfigure}{.49\linewidth}
	\begin{tabular}
	{c@{\hspace{0.1mm}}c@{\hspace{.1mm}}}
        \includegraphics[width=0.5\linewidth]{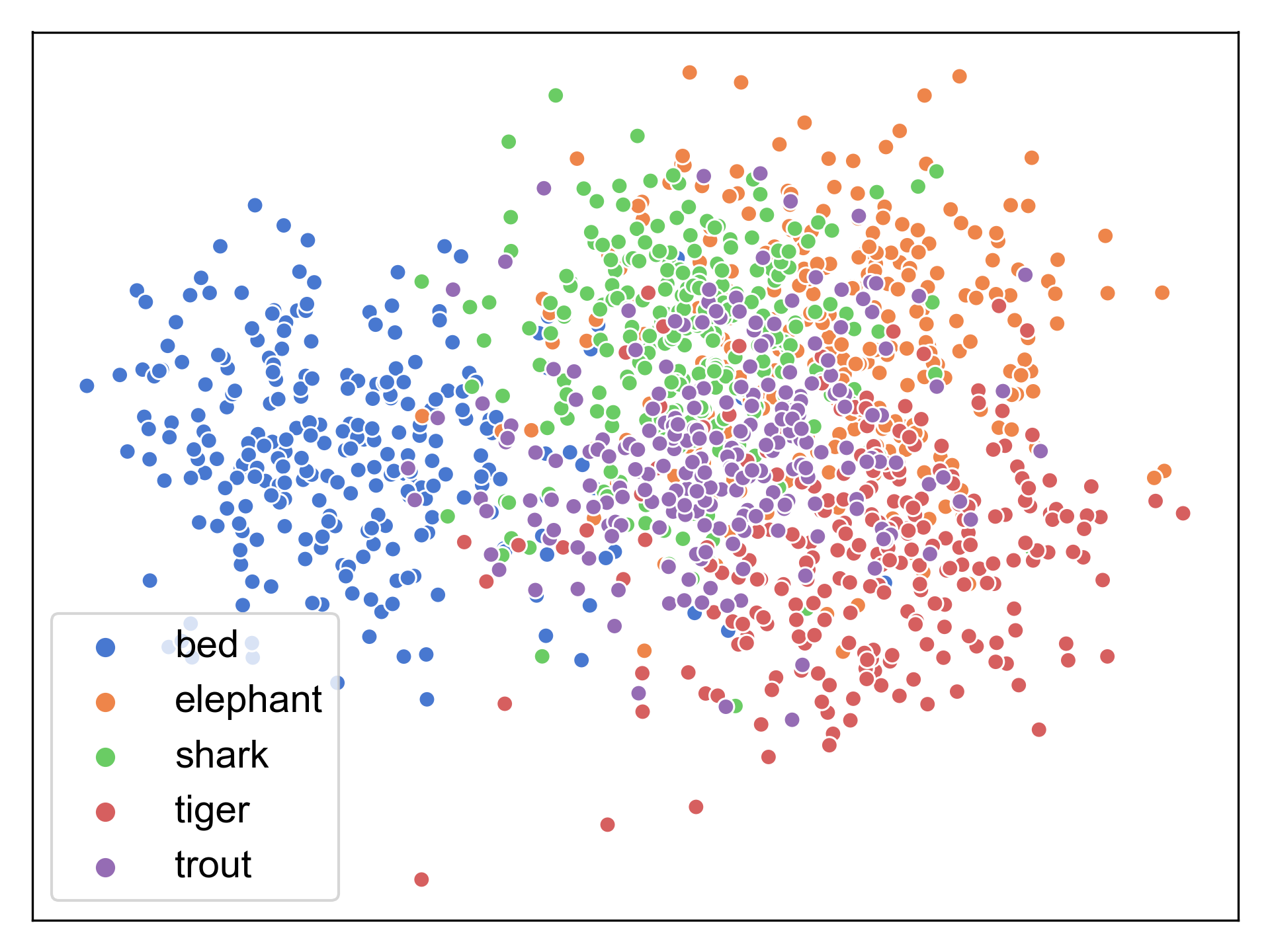} &
    	\includegraphics[width=0.5\linewidth]{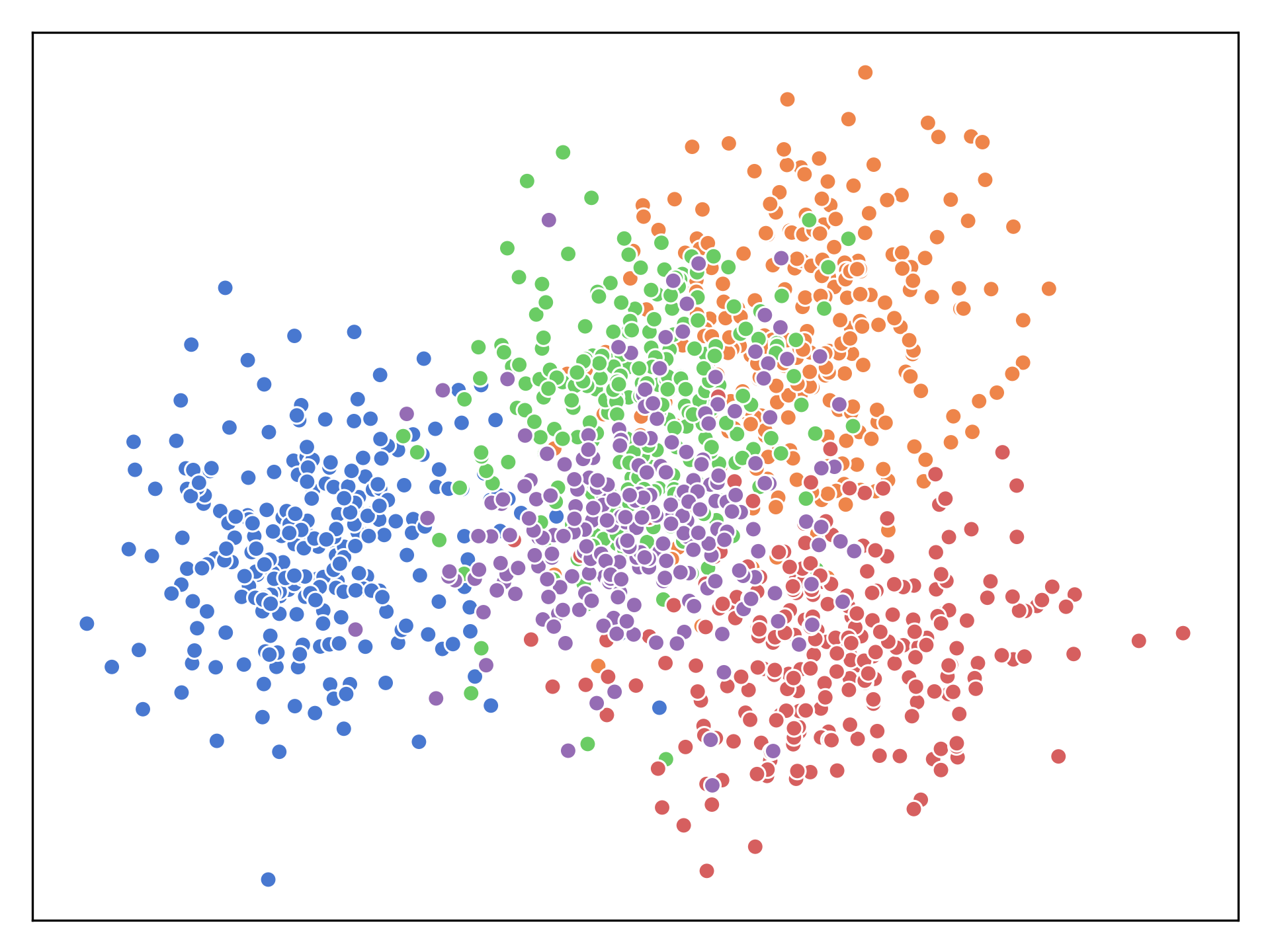}
    	\\
    	\small Class-Generic & \small Class-Specific
	\end{tabular}
    \end{subfigure}
    
    
    \begin{subfigure}{.49\linewidth}
	\begin{tabular}
	{c@{\hspace{0.1mm}}c@{\hspace{.1mm}}}
        \includegraphics[width=0.5\linewidth]{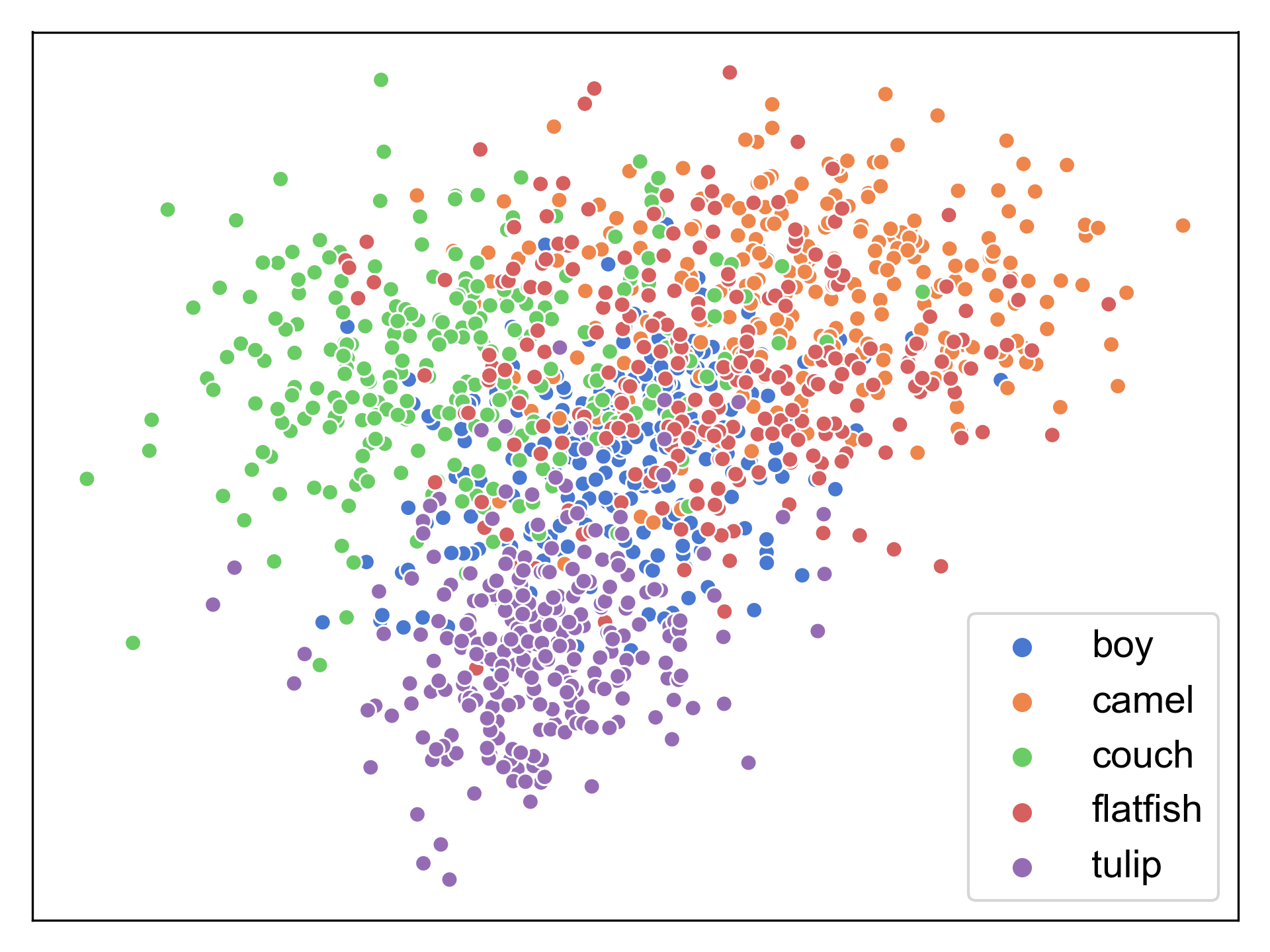} &
    	\includegraphics[width=0.5\linewidth]{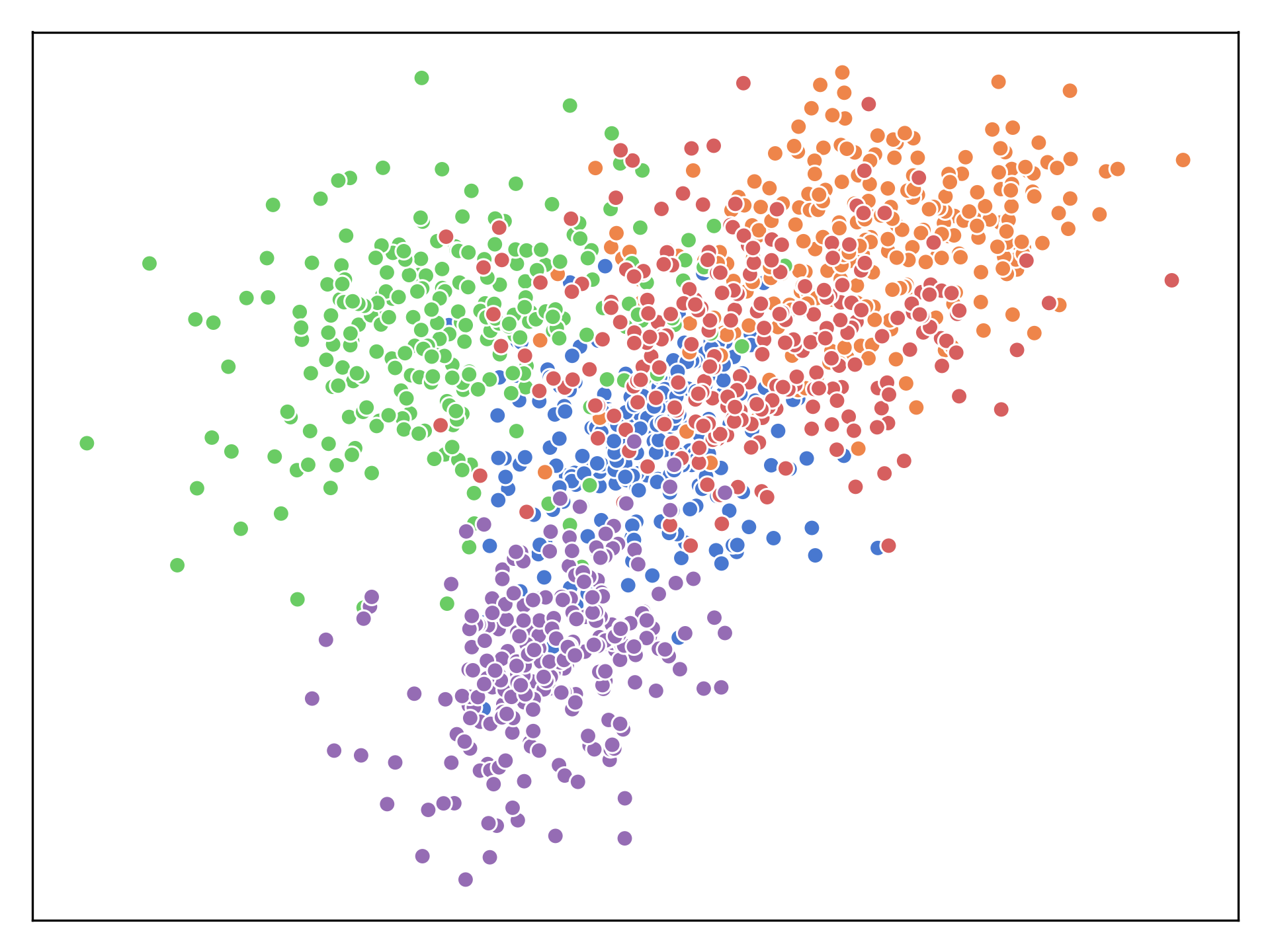}
    	\\
    	\small Class-Generic & \small Class-Specific
	\end{tabular}
    \end{subfigure}
    \hspace{0.1mm}
    \begin{subfigure}{.49\linewidth}
	\begin{tabular}
	{c@{\hspace{0.1mm}}c@{\hspace{.1mm}}}
        \includegraphics[width=0.5\linewidth]{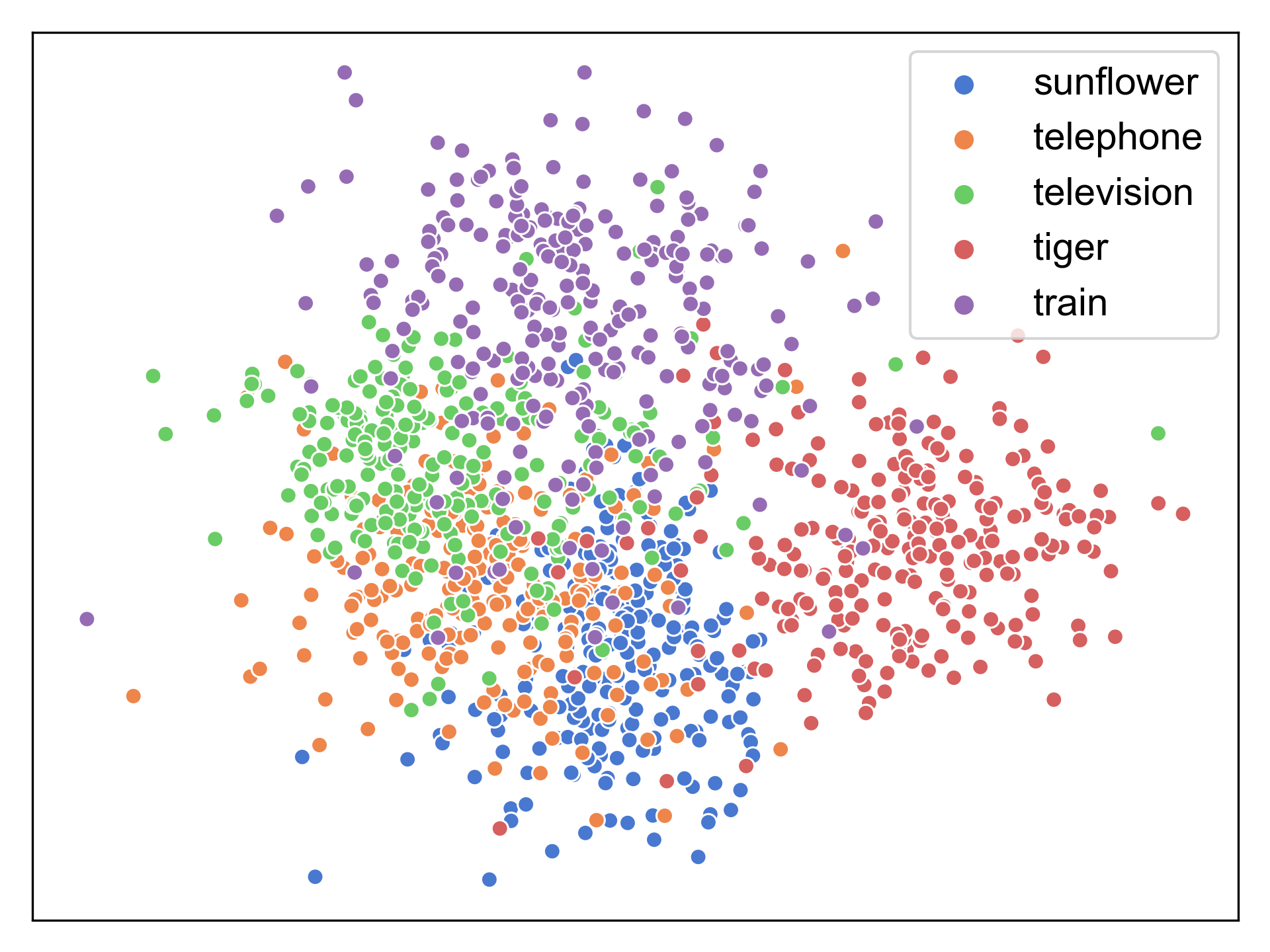} &
    	\includegraphics[width=0.5\linewidth]{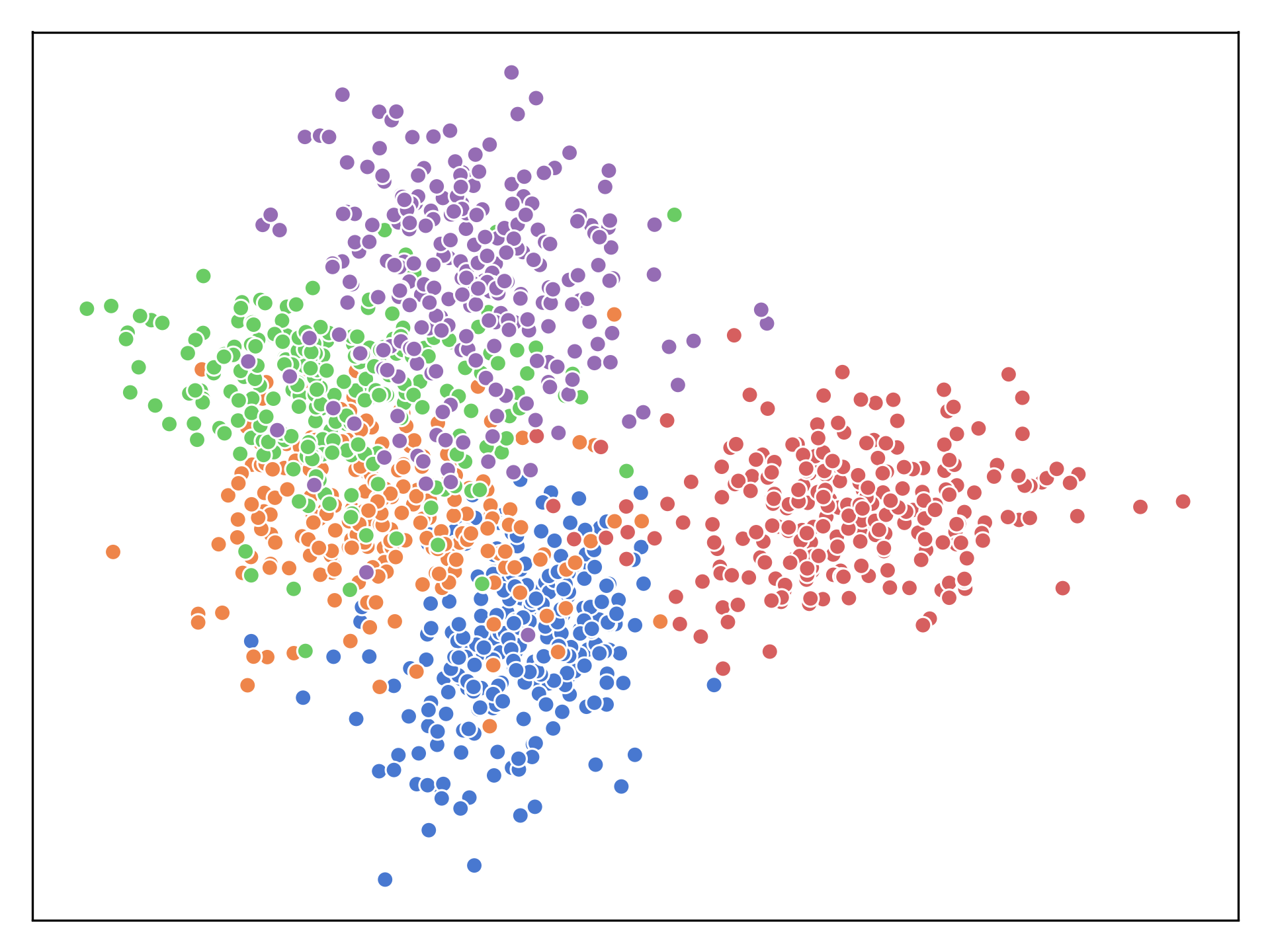}
    	\\
    	\small Class-Generic & \small Class-Specific
	\end{tabular}
    \end{subfigure}
    \vspace{-5px}
    \caption{Feature space visualization for different subset of CIFAR-100 using ResNet-18. }%
    \label{fig:confusion_matrix}%
    \vspace{-3px}
\end{figure}

\end{document}